\documentclass[final,12pt,authoryear]{elsarticle}

\usepackage[margin=2.5cm]{geometry}
\usepackage{graphicx}
\usepackage{subcaption}
\usepackage[space]{grffile}
\usepackage{textcomp}
\usepackage{longtable}
\usepackage{tabulary}
\usepackage{booktabs,array,multirow}
\usepackage{amsfonts,amsmath,amssymb}
\usepackage{siunitx}
\usepackage{parskip}
\usepackage[table, svgnames, dvipsnames]{xcolor}
\definecolor{hnv1}{RGB}{55,166,0}
\definecolor{hnv2}{RGB}{89,184,0}
\definecolor{hnv3}{RGB}{130,205,0}
\definecolor{hnv4}{RGB}{175,222,0}
\definecolor{hnv5}{RGB}{226,243,0}
\definecolor{hnv6}{RGB}{253,223,0}
\definecolor{hnv7}{RGB}{253,169,0}
\definecolor{hnv8}{RGB}{253,113,0}
\definecolor{hnv9}{RGB}{253,54,0}
\definecolor{hnv10}{RGB}{253,0,0}

\usepackage{breakcites}
\usepackage{gensymb}
\usepackage{hyphenat}
\usepackage{makecell, cellspace, caption}
\PassOptionsToPackage{hyphens}{url}
\usepackage{hyperref}
\usepackage[utf8]{inputenc}
\usepackage[english]{babel}
\usepackage{float}

\usepackage{tabularx} 
\usepackage{ragged2e} 

\usepackage{csquotes}

\usepackage{appendix}

\usepackage[final]{changes}

\newcommand{\cellcolorval}[1]{%
    \ifdim #1 pt > 50pt \cellcolor{blue!70}\else%
    \ifdim #1 pt > 40pt \cellcolor{blue!50}\else%
    \ifdim #1 pt > 30pt \cellcolor{blue!30}\else%
    \ifdim #1 pt > 20pt \cellcolor{blue!15}\else%
    \cellcolor{blue!5}\fi\fi\fi\fi%
}

\newif\iflatexml\latexmlfalse

\begin{document}
\begin{frontmatter}

\title{Multimodal classification of forest biodiversity potential from 2D orthophotos and 3D airborne laser scanning point clouds}
\author[1,2]{Simon B.~Jensen\corref{cor1}}
\author[2,3,4]{Stefan Oehmcke}
\author[1,2]{Andreas Møgelmose}
\author[5]{Meysam Madadi}
\author[2,3]{Christian Igel}
\author[1,2,5]{Sergio Escalera}
\author[1,2]{Thomas B.~Moeslund}

\cortext[cor1]{Corresponding author. Email: sbje@create.aau.dk}
\address[1]{Visual Analysis and Perception Laboratory,  Aalborg University, Denmark}
\address[2]{Pioneer Centre for Artificial Intelligence, Denmark}
\address[3]{Department of Computer Science, Copenhagen University, Denmark}
\address[4]{Institute for Visual \& Analytic Computing, Rostock University, Germany}
\address[5]{University of Barcelona and Computer Vision Center, Spain}
\date{}

\journal{Remote Sensing of Environment}

\selectlanguage{english}

\begin{abstract}
Assessment of forest biodiversity is crucial for ecosystem management and conservation. 
While traditional field surveys provide high-quality assessments, they are labor-intensive and spatially limited.
This study investigates whether deep learning-based fusion of close-range sensing data from 2D orthophotos and 3D airborne laser scanning (ALS) point clouds can reliable assess the biodiversity potential of forests.
We introduce the BioVista dataset, comprising \num{44378} paired samples of orthophotos and ALS point clouds from temperate forests in Denmark, designed to explore multimodal fusion approaches.
Using deep neural networks (ResNet for orthophotos and PointVector for ALS point clouds), we investigate each data modality's ability to assess forest biodiversity potential, achieving overall accuracies of 76.7\% and 75.8\%, respectively. 
We explore various 2D and 3D fusion approaches: confidence-based ensembling, feature-level concatenation, and end-to-end training, \added{with the latter} achieving \added{an} overall accuracies of \replaced{82.0\% when separating low- and high potential forest areas.}{80.5\%, 81.4\% and 80.4\%, respectively.}  
Our results demonstrate that spectral information from orthophotos and structural information from ALS point clouds effectively complement each other \replaced{in the assessment of forest biodiversity potential}{in forest biodiversity assessment}.
\end{abstract}



\end{frontmatter}

\section{Introduction}
\label{sec:introduction}

\subsection{Background and motivation}
Forest biodiversity is fundamental to ecosystem functioning, supporting critical processes such as nutrient cycling, carbon sequestration, water regulation, and soil formation, which are vital for planetary health and human well-being~\citep{goldstein2020protecting,thompson2009forest}. 
This essential biodiversity is facing unprecedented threats globally due to accelerating human activities, including deforestation, habitat fragmentation, and climate change~\citep{aerts2011forest,fao-state-of-worlds-forests-2020}. 

The severity of this ongoing loss has spurred significant regional and international initiatives, like the EU Biodiversity Strategy for 2030~\citep{eu-biodiversity-strategy-2020} and the UN's Global Forests Goals and Sustainable Development Goals (SDGs)~\citep{global-forest-goals-2011,sustainable-development-goals-2017}, aimed at increasing protection and enhancing biodiversity. 
Effectively prioritizing conservation efforts and resources under these initiatives requires scalable and accurate methods for identifying areas of high biodiversity value~\citep{kerry-overview-remote-monitoring-biodiversity-conservation-2022}, a challenge this study seeks to address.

\subsection{Traditional assessment methods and limitations}
\added{In general, biodiversity is difficult to quantify, and we often have to resort to measurements of proxy variables that indicate \emph{biodiversity potential}.}
Traditional forest biodiversity assessment methods rely heavily on resource-intensive manual fieldwork. 
Established approaches like the Forest Health Monitoring Program in the United States~\citep{usda-forest-health-monitoring-2022}, the RAINFOR Amazon Forest Inventory Network methodology in South America \citep{malhi-rainfor-2002} and the International Co-operative Program on Assessment and Monitoring of Air Pollution Effects on Forests in Europe \citep{michel-icp-forest-2022} provide valuable qualitative insights through detailed ground surveys.
However, these methods are limited in their spatial and temporal coverage due to the intensive nature of manual tree measurement, plant identification, and habitat assessment protocols. As a result, large-scale biodiversity assessments often rely on extrapolation from sparse sample plots~\citep{kangas-forest-inventories-in-nordic-countris-2018,nord-larsen-nfi-2022}. 

\subsection{Remote sensing and deep learning approaches}
Remote sensing and close-range remote sensing data, such as multispectral images and LiDAR point clouds from airborne sensors \added{(ALS)}, combined with recent advancements in deep learning methods for processing 2D imagery~\citep{kaiming-resnet,dosovitskiy-2021-vistion-transformer,Woo-2023-ConvNeXtV2} and 3D point clouds~\citep{qi-pointnet++,guo-2021-deep-learning-for-3d-point-clouds, deng-pointvector} offers promising capabilities for rapid and large-scale forest assessment~\citep{zhu-2017-deep-learning-in-remote-sensing, yuan2020deep, brandt:24b}.

Recent deep learning approaches have shown to improve performance in tree species classification~\citep{bjerreskov-2021-tree-species} as well as other forest-related variables, such as tree counting~\citep{cheng2024treecounting}, tree height prediction~\citep{li2023treeheightprediction}, forest canopy cover estimation~\citep{liu2023treecover,reiner2023treecover}, tree mortality estimation~\citep{2024-Khatri-Chhetri-tree-mortality}, land cover classification~\citep{kussul2017-landcover-and-crop-classification} directly from various remote sensing imagery.

Similarly, deep learning-based classification of tree species \deleted{classification} \citep{xi-2020-tree-species} and assessment of forest structural characteristics like aboveground biomass~\citep{oehmcke2024biomassregression} and canopy height \deleted{estimation}\citep{lang-2023-canopy-height}, directly from LiDAR point clouds, has seen recent and promising advancements as well.

\replaced{Fusing 2D imagery and 3D LiDAR data can improve  prediction results compared to using the modalities alone. This has, for example, been shown for}{Fusing complementary 2D imagery and 3D LiDAR data has been shown to improve results over using either modality alone for e.g.} land cover classification and above-ground biomass estimation~\citep{2015-jia-land-classification, rastiveis-2015-lidar-aerial-fusion-urban-classification, hang-2020-lidar-hyperspectral-fusion, daneshtalab-2019-cnn-fusion-aerial-lidar, zhang2019deep}. 
But common fusion strategies exhibit notable limitations. 
One widespread approach involves extracting hand-crafted geometric or statistical features from the 3D point clouds to combine with the 2D image data, frequently using traditional machine learning classifiers~\citep{2015-jia-land-classification,rastiveis-2015-lidar-aerial-fusion-urban-classification}. 
\added{Similarly, \citet{rs10020338} combined 2D spectral and 3D structural data through feature engineering and traditional machine learning to assess forest biodiversity potential.}
This manual feature engineering requires domain expertise and can be suboptimal as it risks discarding potentially valuable patterns present in the raw point clouds \added{and images}. Other methods convert or project the 3D LiDAR data into 2D representations (e.g., sparse images, normalized Digital Surface Models, intensity layers) to facilitate fusion, often enabling the use of standard image-based CNNs~\citep{hang-2020-lidar-hyperspectral-fusion, daneshtalab-2019-cnn-fusion-aerial-lidar}.
However, this strategy inevitably sacrifices the rich 3D structural and spatial relationships inherent in the original point cloud data by avoiding direct processing in its native format.

While many forest-related variables have been estimated or assessed, we have not identified studies in the literature that tackle \replaced{the assessment of forest biodiversity potential directly}{forest biodiversity assessment} from 2D imagery, 3D point clouds, or their combination using deep learning or other machine learning techniques.

\begin{figure}[t]
   \centering
   \includegraphics[width=1\textwidth]{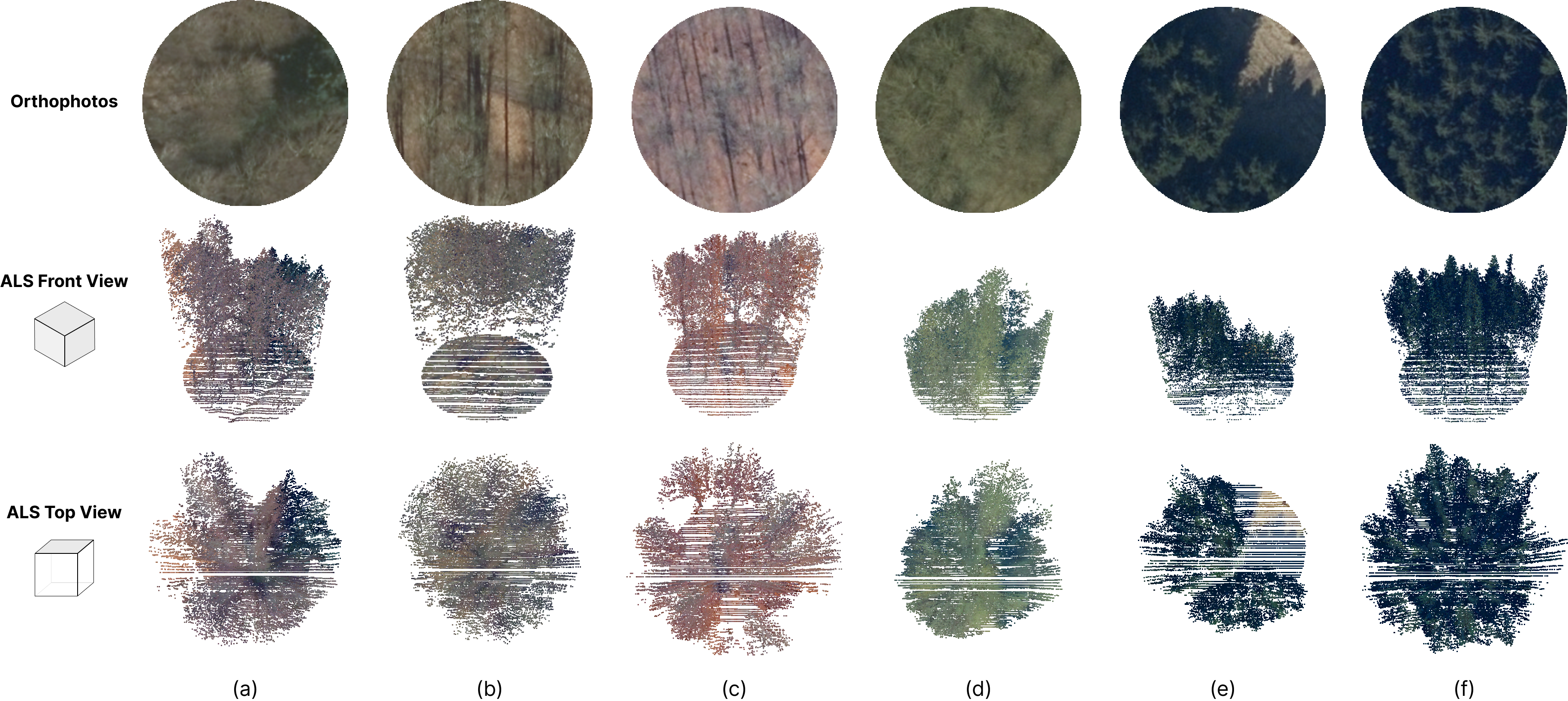}
    \caption{Sample visualization. Paired examples of 2D orthophotos and 3D airborne laser scanning (ALS) point cloud samples from the BioVista dataset, showing both front and top views of point clouds. Examples (a), (b), and (c) represent the high forest biodiversity potential class while (d), (e), and (f) represent the low forest biodiversity potential class.}
   \label{fig:orthophoto-als-pairs}
\end{figure}

\subsection{Research objectives and contributions}
This study investigates how deep learning methods and 2D and 3D close-range remote sensing data, namely 2D orthophotos and 3D ALS point clouds, can be used for assessing forest biodiversity potential.
Furthermore, we explore how the two data modalities complement each other through deep learning multimodal fusion.

Forest biodiversity is an umbrella term which refers to the variety of living organisms such as trees, plants, animals, and microbes found in forest environments~\citep{Kumar2022EcologicalRelevance}. 
In this work, we define forest biodiversity in terms of presence of 11 High Nature Value (HNV) proxy features, introduced by~\citep{johannsen-hnv-skov-2015}, which have been statistically shown to have a high correlation with habitats of endangered species (animals, plants and fungi)~\citep{ejrnaes-2014-biodiversitetskort} within Danish forests. 
\added{Endangered species are great indicators of healthy ecosystems and biodiversity as their presence signifies a stable, resilient, and high-quality ecosystem capable of supporting not only them but a wide array of other, more common species~\citep{ejrnaes-2014-biodiversitetskort}.}
We use the term forest biodiversity potential, since our forest biodiversity definition is based on presences of proxy features, which are indicators of biodiversity potential, rather than being a direct inventory of biodiversity.
Further details on the HNV proxy features are provided in Section~\ref{subsec:biodiversity-ground-truths} and the list of the 11 features are presented in Table~\ref{tab:hnv-proxy-feature-descriptions}.

We introduce the BioVista dataset - a comprehensive collection of paired orthophotos and airborne laser scanning (ALS) point clouds from temperate forests in Denmark, labeled as high- or low forest biodiversity potential.
Examples of orthophotos and ALS point cloud pairs from the dataset are seen in Figure~\ref{fig:orthophoto-als-pairs}.
  
We summarize the contributions of this work as follows:
\begin{itemize}
    \item \textbf{The BioVista dataset}: \replaced{The BioVista dataset was developed,}{Development the BioVista dataset,} containing \num{44378} pairs of 2D orthophotos and 3D ALS point clouds, labeled as high- or low forest biodiversity potential. 
    This dataset and the associated annotations are made publicly available and provide a valuable resource for close-range remote sensing-based \replaced{assessment of forest biodiversity potential}{forest biodiversity assessment}.
    \item \textbf{Deep learning-based forest biodiversity assessment}: We present the first study which assesses forest biodiversity directly from 2D orthophotos and 3D ALS point clouds using deep learning-based methods.
    \item \textbf{Multimodal fusion}: We evaluate deep learning-based multimodal fusion methods that combine 2D spectral- and 3D structural data and clearly demonstrate how the two data modalities complement each other effectively for  \replaced{assessment of forest biodiversity potential}{forest biodiversity assessment}. 
\end{itemize}

The insights and methodologies developed through this research contribute to the broader goal of developing efficient, scalable approaches for forest biodiversity assessment and monitoring using close-range sensing technologies.

The remainder of this paper is organized as follows: 
Section~\ref{sec:dataset} presents the BioVista dataset, including data collection, preprocessing and preparation. 
Section~\ref{sec:method} details our deep learning methodology, covering both single-modality approaches (2D orthophoto and 3D ALS point cloud classification) and multimodal fusion strategies. Section~\ref{sec:experiments} presents our experimental results and performance analysis.
Section~\ref{sec:discussion} discusses limitations and potential improvements of our approach \replaced{before we conclude in Section~\ref{sec:conclusion}}{and Section~\ref{sec:conclusion} concludes the work with key findings and future research directions}.

\section{Dataset}
\label{sec:dataset}

\begin{figure}[t]
   \centering
   \includegraphics[width=1\textwidth]{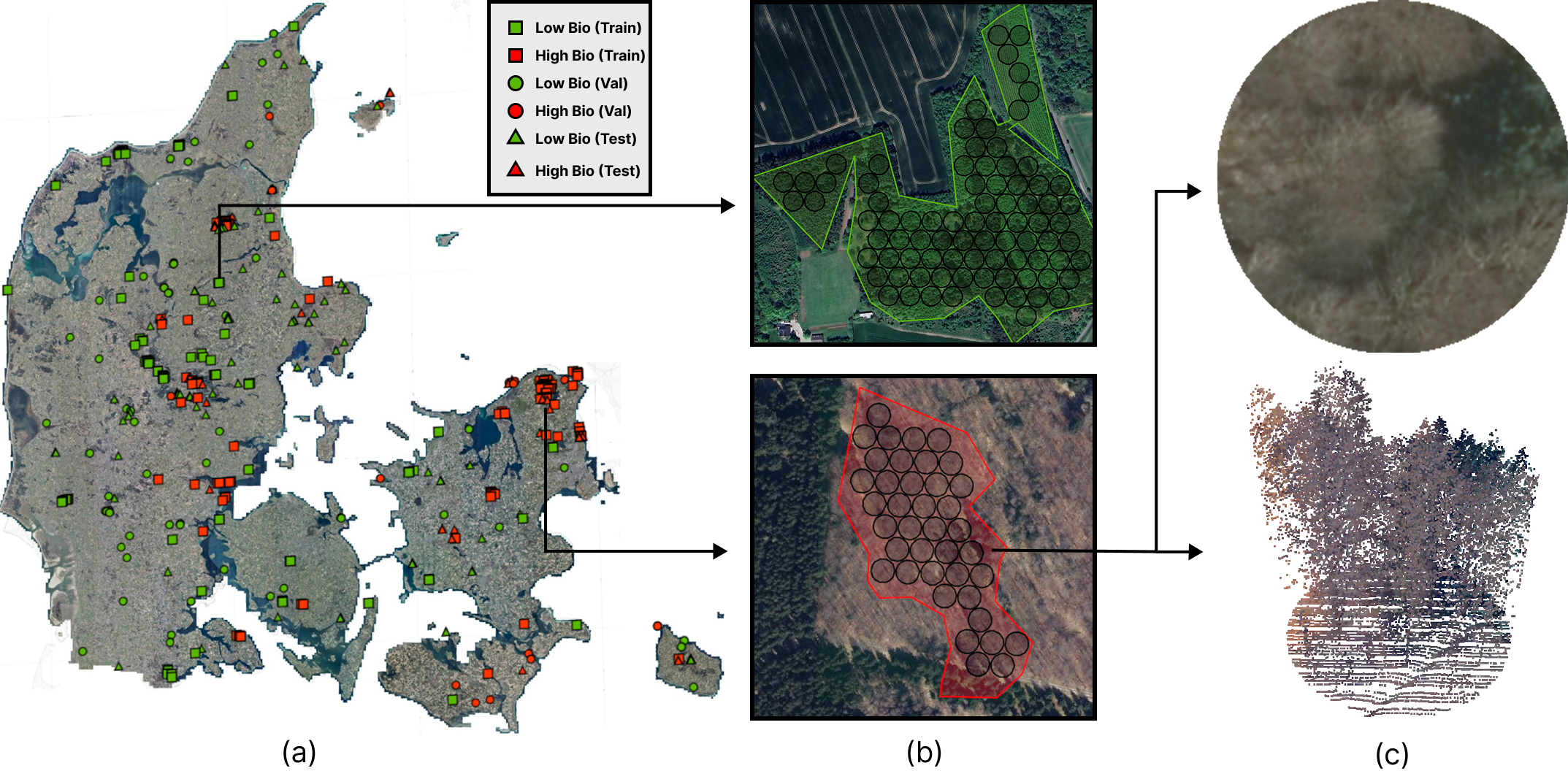}
   \caption{The BioVista dataset. (a) Geographic distribution of forest areas used for sampling training, validation and test sample plots\added{, shown on the orthophoto map of Denmark}. 
   (b) Example of two forest areas which contain 84 and 39 circular samples. Samples have a diameter of 30 meters and samples from the same forest area belong to the same class. 
     (c) A single sample is represented as a 2D orthophoto, where 1 pixel is equal to 12,5 cm on the ground, and a 3D ALS point cloud with 8 points/m².}
   \label{fig:biovista-dataset-structure}
\end{figure}

\subsection{The BioVista dataset}
The BioVista dataset, introduced in this study, combines High Nature Value (HNV) proxies \citep{johannsen-hnv-skov-2015}, described in detail in Section~\ref{subsec:biodiversity-ground-truths}, with 2D orthophotos and 3D ALS point clouds to create a comprehensive resource for forest biodiversity potential classification. 
The dataset is a unique, high-quality resource for forest biodiversity \replaced{potential}{assessment}, made possible by Denmark’s tradition of comprehensive forest registries \citep{nord-larsen-nfi-2022}, extensive geographic data collection, and open data policies \citep{hansen-e-governance-2011}.

\added{Forests cover about 636,079 ha of Denmark, which corresponds to  a forest fraction of 14.8\%, according to  \cite{bjerreskov-2021-tree-species}. The fractions of conifer and broadleaf forests were estimated to be 45\% and 55\% of the forest area, respectively \citep{bjerreskov-2021-tree-species}.
The most common broadleaf trees in Denmark are (in order)
beech, oak, birch, Sycamore maple, and ash; the most common
conifers are Norway spruce, pine species, Sitka spruce, Nordmann fir, and Noble fir \citep{nord-larsen-nfi-2022}.
Several tree height maps of Denmark have been made available recently, we refer to \cite{liu2023treecover} and \cite{li2023treeheightprediction}.}

The  dataset comprises \num{44378} pairs of 2D orthophoto images and 3D ALS point clouds derived from forests in Denmark, labeled as either high- or low biodiversity potential. 
An overview of the dataset, geographic location of forest areas and sampling methodology can be seen in Figure~\ref{fig:biovista-dataset-structure}.
This shape and size of samples conforms to the test plot standard used in, e.g., the National Forest Inventory (NFI) of Denmark~\citep{alban-nfi-field-instructions}, and is comparable to other nordic NFI standards~\citep{kangas-forest-inventories-in-nordic-countris-2018}. 
One sample covers an area of $\sim707$ m$^2$ and the rational is to have a large enough area to include a diverse selection of both small and larger trees and vegetation within a single sample, as well as variations in soil, ground cover, and other ecological factors within the same area.

\subsection{Biodiversity \replaced{potential labels}{ground truths}}
\label{subsec:biodiversity-ground-truths}

\begin{figure}[t]
   \centering
   \includegraphics[width=1\textwidth]{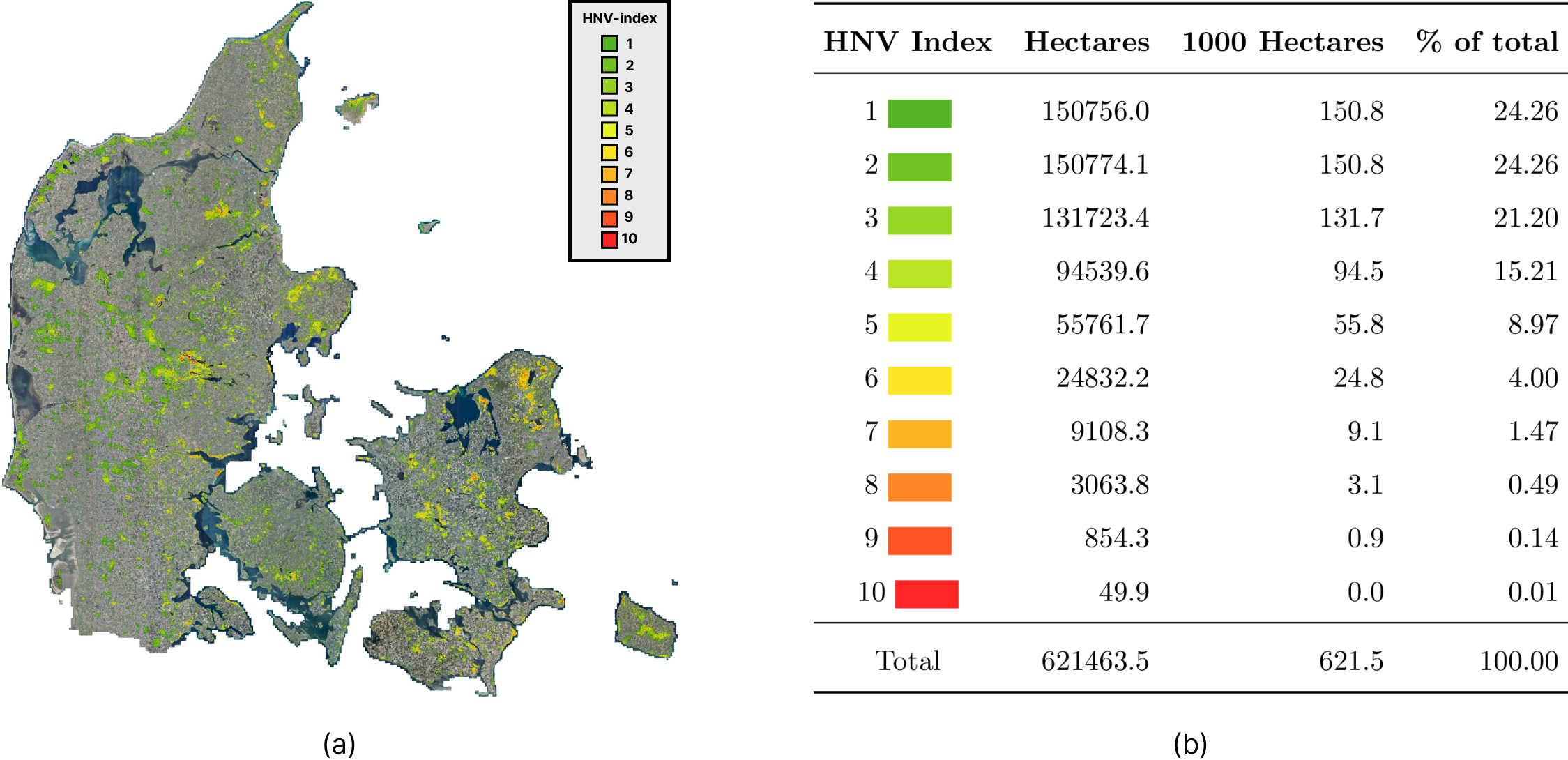}
   \caption{HNV forest distribution. (a) High Nature Value (HNV) forest map of Denmark~\citep{johannsen-hnv-skov-2015}, which shows the spatial distribution of forest areas with HNV scores from 1 to 10 where 1 is worst and 10 is best \added{, shown on the orthophoto map of Denmark}. (b) Quantitative distribution showing that forest areas with HNV scores from 1-3 comprise nearly 75\% of Danish forests, while areas with scores from 7-10 represent approx. 2\%.}
   \label{fig:denmark-hnv-index}
\end{figure}

We use the HNV forest map~\citep{johannsen-hnv-skov-2015}, seen in Figure~\ref{fig:denmark-hnv-index}, as the foundation for gathering ground truth labels for our samples. 
The HNV forest map is based on 11 features that serve as proxies for forest biodiversity, combining historical and current data.
Proxy features include, among others, canopy height variation, coastal proximity, tree and plant richness, forest cover, occurrence of large trees, and occurrence of inner forest edges.
A short description of each proxy feature is shown in Table~\ref{tab:hnv-proxy-feature-descriptions}, but we refer to \added{\cite{johannsen-hnv-skov-2015} for details.}

The HNV score for each forest area is calculated by summing the number of present proxy features. 
A score is incremented by 1 for each feature present, resulting in a score between 0 and 11.
However, no forest area achieves a score of 11.
Figure~\ref{fig:denmark-hnv-index} shows the quantitative distribution for each score.
Although the color scheme may appear counterintuitive, with green representing lower biodiversity values and red representing higher ones, we choose to maintain consistency with the original HNV forest map convention. 

\begin{table}[H]
\centering 
\footnotesize 
\begin{tabularx}{\textwidth}{@{} l >{\RaggedRight}X @{}} 
\toprule
\textbf{Proxy Feature} & \textbf{Description} \\
\midrule
Coastal Proximity & Areas within 1 km of the coastline, often less cultivated and with dynamic natural processes supporting biodiversity. \\
\midrule
Height Variation & Variation in tree height/age within 25$\times$25m plots (LiDAR based), indicating structural diversity and more habitats. \\
\midrule
Woody Plant Richness & Variety of woody plant species in 4$\times$4 km cells (NFI based), suggesting diverse conditions and less intensive forestry. \\
\midrule
Large Trees & Presence of trees $>60$ cm diameter at breast height in 4$\times$4 km cells (NFI based), indicating continuity and providing specific habitats. \\\midrule
Forest Structure & Combines habitat mapping and fungi data to identify areas with features like large/old trees, dead wood, and extensive management. \\\midrule
Forest Continuity & Areas where current forest overlaps with forest mapped ca. 1780, indicating higher probability of long-term continuity. \\\midrule
Nature Share 40\% & Landscape context where at least 40\% of the surrounding 1$\times$1 km area is natural habitat, important for species connectivity. \\\midrule
Nature Share 80\% & Landscape context where at least 80\% of the surrounding 1$\times$1 km area is natural habitat, indicating very high connectivity. \\\midrule
Inner Forest Edge & Forest areas within 120m of large open areas inside the forest, offering habitat variation and transition zones. \\\midrule
Habitat Nature & Areas mapped as EU Habitats Directive types, likely having high nature value due to focus on intact nature. \\\midrule
Mapped Nature & Combines §3 nature areas, valuable §25 forests, oak scrub, and state-designated natural forests, based on known biodiversity value. \\
\bottomrule
\end{tabularx}
\caption{HNV forest proxy features. A short summary description of each of the 11 proxy features of the HNV forest map. we refer to~\citep{johannsen-hnv-skov-2015} for an exhaustive description.} 
\label{tab:hnv-proxy-feature-descriptions}
\end{table}

\added{In pilot experiments, we realized that it is very difficult to predict the exact HNV scores from remote sensing data. Thus, we simplified the task by distinguishing three aggregated classes, low, medium, and high biodiversity potential, each of which comprising roughly the same number of different HNV score values.}
We define low forest biodiversity potential as areas with HNV scores from 1--3, high forest biodiversity potential as areas with HNV scores from 7--10, \replaced{and medium forest biodiversity potential as areas with HNV scores from 4--6.}{while ignoring forest with HNV scores between 4 and 6.} 
\added{The focus of this study was on the binary task of distinguishing low and high biodiversity potential areas. Arguably, this is an important subproblem, as one wants to avoid mixing up these two classes in forest management decisions.}
\added{Starting with simpler coarse-grained tasks that can be solved with an accuracy high enough to be regarded as informative} and allows for a \replaced{meaningful}{more focused} comparison of deep learning-based classification methods trained on 2D orthophotos, 3D ALS point clouds, and a combination of the two, while still evaluating the forest biodiversity assessment capabilities of deep learning-based methods. 

\subsection{2D orthophotos}
\label{subsec:2d-orthophotos}
The 2D orthophotos used in this project are derived from yearly nationwide aerial photography of Denmark and are part of GeoDanmark's geographic administration framework~\citep{geodanmarkforaarsbilleder}. 
Orthophotos are created by adjusting each pixel according to the camera's position and the elevation model of the Earth's surface, ensuring that the images are geometrically corrected and can be used for precise measurements.

\begin{figure}[t]
   \centering
   \includegraphics[width=1.0\textwidth]{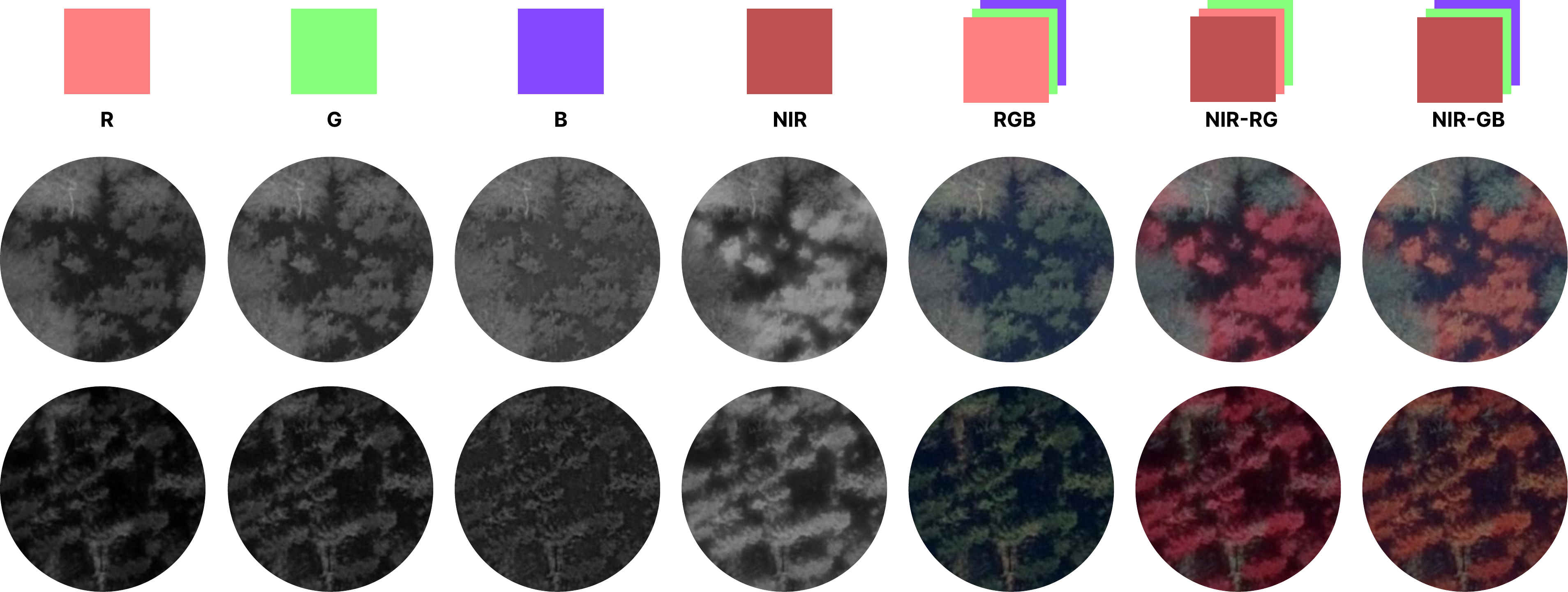}
   \caption{Spectral information in orthophotos.
   Two samples from the BioVista dataset represented in various combinations of color bands of red, green, blue, near-infrared.}
   \label{fig:color-channels-in-orthophotos}
\end{figure}

The orthophotos have a ground resolution of 12.5 cm per pixel. 
They contain spectral information across the visible light spectrum, comprising Red (R), Green (G), and Blue (B) channels spanning approximately 380--740 nm and include a Near-infrared (NIR) channel spanning approximately 750--1400 nm.
NIR is particularly interesting in terms of forest and vegetation assessment, because healthy vegetation strongly reflects NIR light.
This characteristic is illustrated in Figure~\ref{fig:color-channels-in-orthophotos}, which shows two forest samples from the BioVista dataset represented using individual R, G, B, and NIR channels, as well as standard RGB and two false-color composites (NIR-RG and NIR-GB).
The figure demonstrates the high intensity signal from vegetation within the NIR channel specifically.
The orthophotos are captured annually before leaf-out, between March 1st and May 1st and the images undergo extensive post-processing to ensure consistent colors, contrast, and brightness across the capture period.
The resulting map of Danmark, from the 2023 orthophoto collection, can be seen in Figure~\ref{fig:orthophotos-denmark-with-ortho-and-als-examples}.

\subsection{3D ALS point clouds}
\label{subsec:3d-als-point-clouds}
\begin{figure}[t]
   \centering
   \includegraphics[width=0.6\textwidth]{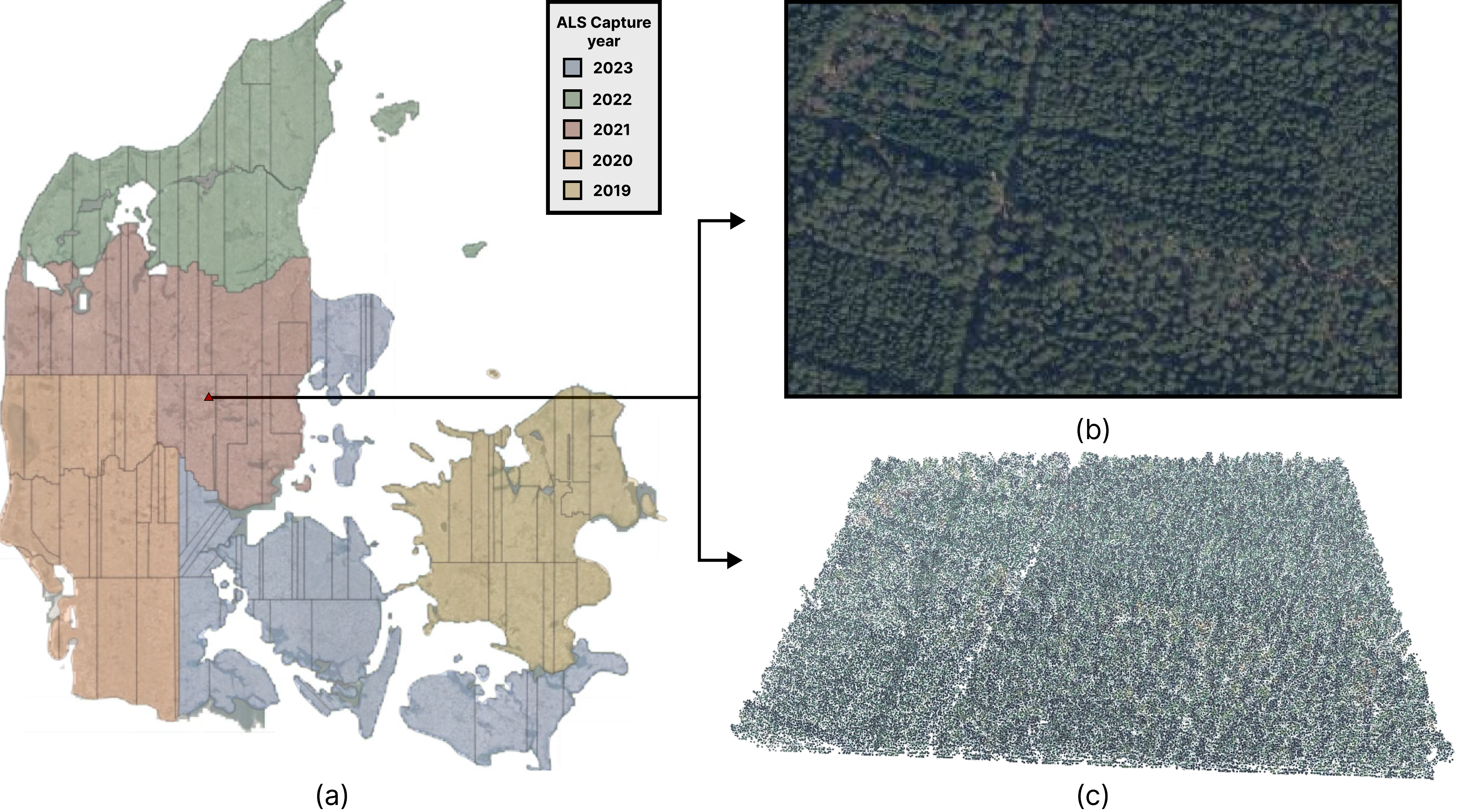}
   \caption{Remote sensing data sources. Orthophotos and ALS point cloud coverage of Denmark. 
   (a) 2023 orthophoto map with ALS capture years overlaid. 
   (b) High-resolution orthophoto example (56.161144°N, 9.594725°E). 
   (c) Corresponding ALS point cloud.}
   \label{fig:orthophotos-denmark-with-ortho-and-als-examples}
\end{figure}

The 3D ALS point clouds used in this study is part of the Danish Height Model dataset~\citep{dhmklimadatastyrelsen}. \added{The ALS product specifications can be found in the report by the
\cite{dhm2020}.}
This comprehensive dataset covers the entire country of Denmark and is continuously updated through yearly systematic airborne laser scanning.
As of 2018 and forward, 1/5 of the country is updated annually.
In this research we use ALS point clouds from 2019, 2020, 2021, 2022 and 2023 depending on the area of the forest, as illustrated in Figure~\ref{fig:orthophotos-denmark-with-ortho-and-als-examples}.
For each forest area, we ensure temporal consistency by using orthophotos captured in the same year as the corresponding ALS point cloud data, allowing for fair comparison between the two modalities and ensuring that both data sources represent the forest conditions from the same time period.
The point clouds are available in tiles spanning 1$\times$1 km and the point density averages 8 points/m$^2$ for data, with a horizontal accuracy of 0.15 m and a vertical accuracy of 0.05 m. 

\subsection{Dataset creation process} \label{sec:dateset_creation}

The creation process of the BioVista dataset involves the following steps:
\begin{enumerate}
    \item \replaced{Partitioning}{Filtering} the HNV forest map (seen in Figure~\ref{fig:denmark-hnv-index}) \replaced{into}{to include only forest} areas with scores from 1--3, 
    7--10, \replaced{representing}{used for extracting} low
 and high forest biodiversity potential samples, respectively.
    \item Filtering narrow forest areas using morphological opening and exclude forest areas with an area below 0.5 hectares using connected components analysis, thereby following the minimum forest size definition of~\cite{fao-state-of-worlds-forests-2020}.
    \item Randomly selecting \replaced{forest areas for each biodiversity class from the filtered areas to avoid class-imbalance, whereby we chose 130 areas per class.}{130 low forest biodiversity potential areas and 130 high forest biodiversity potential areas from the filtered areas.}
    \item Subdivide the initial forest areas, by excluding parts which contains roads, forest track or tree-less open areas, to ensure the areas consists purely of forest covered areas. Resulting in 145 high- and 231 low forest biodiversity potential areas. See Figure~\ref{fig:biovista-dataset-structure} (a) for an overview of the forest areas.
    \item Extracting non-overlapping coordinates for samples from each forest area, as illustrated in Figure~\ref{fig:biovista-dataset-structure} (b).
    \item Downloading 2D orthophotos and 3D ALS point clouds for the sample coordinates, see Figure~\ref{fig:biovista-dataset-structure} (c). 
    We ensure temporal consistency between downloaded orthophotos and ALS point clouds, meaning they have the same capture year.
\end{enumerate}

\begin{table}[th]
\footnotesize
\caption{Dataset distribution. 
Number of samples and areas across the training-, validation- and test sets. In addition, statistics on the average number of samples per area and average area size in hectares for both the high- and low forest biodiversity potential class.}
\label{tab:data-split}
\centering
\begin{tabular}{llrrrr}
\toprule
 & \multicolumn{1}{c}{\textbf{Category}} & \multicolumn{1}{c}{\textbf{\# Areas}} & \multicolumn{1}{c}{\textbf{\# Samples}} & \multicolumn{1}{c}{\textbf{Avg. samples}} & \multicolumn{1}{c}{\textbf{Avg. size (ha)}} \\ 
 \midrule
\multirow{2}{*}{Train}  & High Bio. Potential & 88 & \num{15507} (34.9\%)   & 176.2  & 19.9   \\  
      & Low Bio. Potential  & 95 & \num{11418} (25.7\%)   & 120.2  & 13.8   \\
\addlinespace
\multirow{2}{*}{Val} & High Bio. Potential & 16 & \num{4116} (9.3\%)  & 257.3   & 28.7 \\  
    & Low Bio. Potential  & 58 & \num{4604} (10.4\%)  & 79.4  & 10.1 \\ 
\addlinespace
\multirow{2}{*}{Test}     & High Bio. Potential & 41 & \num{4287} (9.7\%)  & 104.6  & 12.2 \\  
        & Low Bio. Potential  & 78 & \num{4446}  (10.0\%) & 57.0 &  7.6 \\ 
\midrule
\textbf{Total} & & 376 & \num{44378} & 118.0 & 16.1 \\
\bottomrule
\end{tabular}
\end{table}

\subsection{Dataset split}
\begin{figure}[t]
   \centering
   \includegraphics[width=1\textwidth]{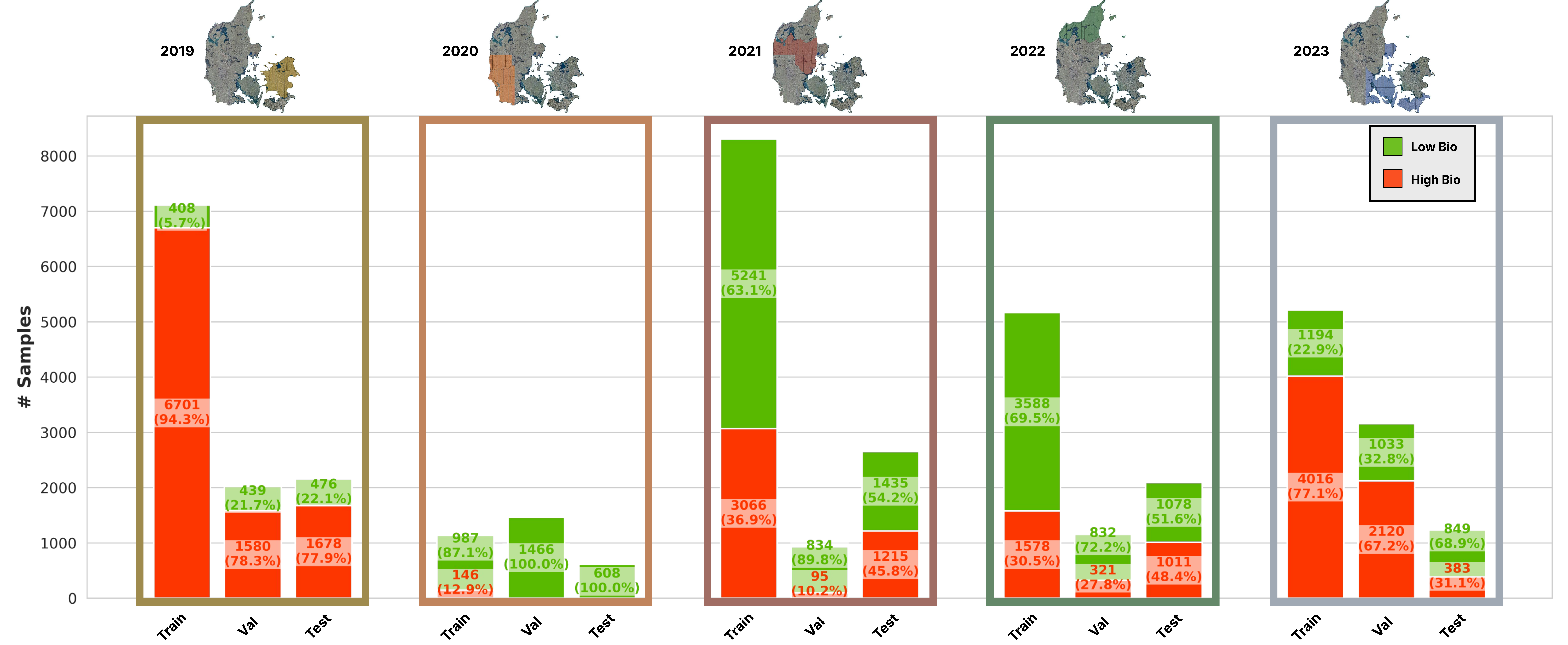}
   \caption{Regional dataset distribution. 
   Bar plot showing the number of high- and low forest biodiversity potential samples across the train, validation and test sets per region.}
   \label{fig:regional-dataset-split-per-class-bar-plot}
\end{figure}

The 145 high- and 231 low forest biodiversity potential areas are randomly split into the training, validation or test sets, using a 60\%-20\%-20\% split ratio, resulting an overall dataset distribution as seen in Table~\ref{tab:data-split} and regional dataset distribution as seen in Figure~\ref{fig:regional-dataset-split-per-class-bar-plot}.
When splitting the forest areas, we incorporate a spatial constraint that ensures forest areas which are split into the training and validation sets have a minimum distance of 5 km from each other to reduce risk of spatial autocorrelation.
The min- and mean distances between areas in the training set and the nearest area in the validation set is 4.93 km and 16.48 km.
This constraint is crucial as the validation set is used during hyperparameter tuning and model selection during our experiments. 
We imposed no spatial constraint between areas in the training set and test set, where the min- and mean distances by nearest areas are 0.41 km and 12.43 km, nor between areas in the validation set and test set where min- and mean distances by nearest areas are 0.36 km and 15.33 km. 
The location of training-, validation- and test areas can be seen in Figure~\ref{fig:biovista-dataset-structure}.

\section{Method}
\label{sec:method}
\begin{figure}[!t]
   \centering
   \includegraphics[width=0.80\textwidth]{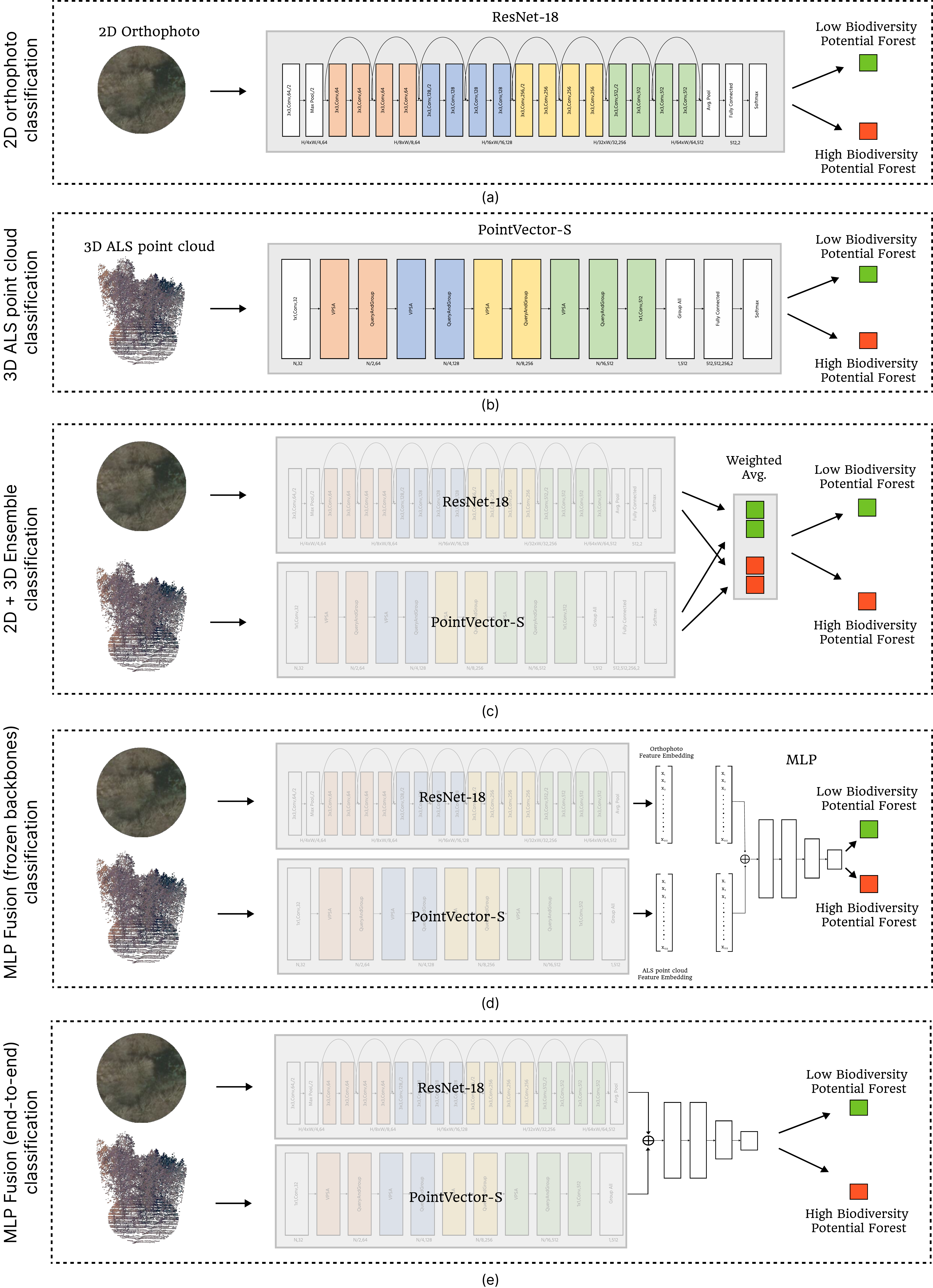}
   \caption{Classification method overview. 
   Architectural overview of the methods used for classifying samples into high- and low forest biodiversity potential.
   (a) 2D orthophoto classification using ResNet-18, 
   (b) 3D ALS point cloud classification using PointVector-S, 
   (c) Ensemble classification combining prediction probabilities through weighted averaging, 
   (d) MLP Fusion trained \replaced{with frozen backbones}{on pre-computed feature embeddings from both modalities} and 
   (e) MLP Fusion trained end-to-end.}
   \label{fig:method-overview}
\end{figure}

In this section, we describe our methodology for assessing the forest biodiversity potential of samples, using various deep learning methods. 
We explore five distinct methods: 2D orthophoto classification using ResNet~\citep{kaiming-resnet}, 3D ALS point cloud classification using PointVector \citep{deng-pointvector} and three multimodal fusion methods that combine the 2D- and 3D modalities. 
An architectural overview of all methods are shown in Figure~\ref{fig:method-overview}.

\subsection{2D orthophoto image classification}
\label{subsec:2d-orthophoto-image-classification}

For the classification of forest biodiversity potential of samples using strictly the 2D orthophotos, we employ the ResNet architecture~\citep{kaiming-resnet}.
ResNet is a deep convolutional neural network (CNN) architecture known for its ability to train effectively by using residual connections to mitigate the vanishing gradient problem. 
The architecture is well-suited for image classification tasks where the model must learn complex patterns from large-scale image datasets.
A ResNet consists of a combination of convolutional layers, batch normalization, and ReLU activation functions, followed by a fully connected layer and a softmax layer, as visualized in Figure~\ref{fig:method-overview} (a).
While newer architectures for 2D image classification, such as the Vision Transformer (ViT)~\citep{dosovitskiy-2021-vistion-transformer} and ConvNeXt~\citep{Woo-2023-ConvNeXtV2}, demonstrate state-of-the-art performance on many benchmarks, ResNet still remains a valid choice, due to its proven performance, computational efficiency, ease of use, and well-understood behavior.
We refer to the original work for further details on the architecture. 

\subsection{3D ALS point cloud classification}
\label{subsec:3d-als-point-cloud-classification}
For the task of classifying the forest biodiversity potential of samples based on 3D ALS point clouds, we employ the PointVector architecture~\citep{deng-pointvector}.
PointVector builds upon previous successful point-based networks such as PointNet, PointNet++~\citep{qi-pointnet++} and PointNeXt~\citep{guocheng-pointnext}. 
PointVector is a state-of-the-art point cloud processing model that introduces a novel vector-oriented point set abstraction module to enhance local feature aggregation, allowing for more effective extraction of spatial information from point cloud data.
The model has demonstrated \replaced{state of the art}{SOTA} performance on popular 3D point cloud benchmarks such as S3DIS~\citep{armeni-s3dis} and ScanObjectNN~\citep{uy-scanobjectnn}, while using significantly fewer parameters than comparable models~\citep{opentrench3d}.
An architectual overview of PointVector is shown in Figure~\ref{fig:method-overview} (b).
PointVector extracts features directly from raw 3D ALS point clouds without manual feature engineering. However, the architecture requires a fixed number of input points (e.g., \num{16384}). 
Since the ALS point clouds in the BioVista dataset often exceed \num{30000} points, down-sampling is necessary, presenting a limitation as potentially relevant spatial information from discarded points is neglected.
Further details on the PointVector architecture and its inner workings are beyond the scope of this paper and can be found in the original work.

\subsection{2D-3D multimodal classification}
One challenge in combining 2D orthophotos and 3D ALS point clouds for classification is their fundamentally different data structures.
Images are regular, ordered and discrete while point clouds are irregular, order-less and continuous. 
These differences has led to different feature extraction methodologies, as presented in Section~\ref{subsec:2d-orthophoto-image-classification} and~\ref{subsec:3d-als-point-cloud-classification} and shown in Figure~\ref{fig:method-overview}. 
To address this challenge and leverage the complementary information from both modalities, we explore three distinct fusion approaches. 

\subsubsection{Confidence-based ensembling}
Our first fusion method, visualized in Figure~\ref{fig:method-overview} (c), combines the predictions from the individual models already trained: the ResNet model (for 2D orthophotos) and the PointVector model (for 3D point clouds). 
This technique is a form of confidence-based ensembling/stacking \citep{wolpert1992stacked}.

The process works as follows: For each input sample (which consists of a paired 2D orthophoto and 3D ALS point cloud for a specific forest sample), we first get the predictions from both models independently. 
Each model outputs a probability score indicating its confidence that the input sample belongs to \replaced{a particular}{either the high- or low} forest biodiversity potential class.

We then combine these probabilities using a weighted average to get a final score for each class:
\[
P(\text{class}\,|\,\text{input} ) = w_{\text{2D}} \cdot P_{\text{2D}}(\text{class}\,|\,\text{input} ) + w_{\text{3D}} \cdot P_{\text{3D}}(\text{class}\,|\,\text{input})\enspace,
\]
where $P_{\text{2D}}$ and $P_{\text{3D}}$ are the class probabilities from the ResNet and PointVector models respectively, and $w_{\text{2D}}$ and $w_{\text{3D}}$ are the corresponding weights. 
These weights are pre-determined by finding the weight values that yield the best classification performance on our separate BioVista validation dataset.
This approach allow each model to contribute according to its relative strength in classification, while maintaining the independence of the 2D- and 3D processing pipelines.

\subsubsection{MLP fusion}
We explore two distinct Multi-Layer Perceptron (MLP) based fusion strategies to integrate information from the 2D and 3D domains, as illustrated in Figure~\ref{fig:method-overview} (d) and (e). 
Both approaches aim to combine the spectral information captured by ResNet-18's image features with the structural information encoded in PointVector-S's point cloud features.

The core process shared by both methods involves:
\begin{itemize}
\item Extracting 512-dimensional feature vectors from the respective backbones (ResNet-18 for images, PointVector-S for point clouds).
\item Concatenating these vectors to form a unified 1024-dimensional feature representation.
\item Feeding this combined vector into an MLP classifier for the final prediction.
\end{itemize}

The MLP architecture used in both cases consists of an input layer matching the 1024 concatenated features, followed by two hidden layers with sizes 512 and 256 respectively. 
ReLU activation and batch normalization are applied after each hidden layer.
The two MLP fusion methods differ significantly in their training paradigm:

\textbf{MLP Fusion (\replaced{frozen backbones}{pre-computed})}:
In this approach, illustrated in Figure~\ref{fig:method-overview} (d), the ResNet-18 and PointVector-S backbones act solely as \replaced{frozen (non-trainable) high-dimensional}{fixed} feature extractors. 
\replaced{During training and evaluation, input samples are passed through these backbones to generate 512-dimensional feature embeddings, which are then fed into the MLP classifier.
Only the MLP parameters are updated, while the backbone weights remain fixed throughout the process.
}{
We first process the entire BioVista training, validation and test datasets through the frozen (non-trainable) backbones to generate and store the 512-dimensional feature embeddings for every sample. 
Subsequently, only the MLP classifier is trained, validated, and evaluated using these pre-computed, static feature vectors. 
The backbones themselves are not active or updated during this MLP training phase.}

\textbf{MLP Fusion (end-to-end)}: This method represents a more integrated training process, as illustrated in Figure~\ref{fig:method-overview} (e). 
Both the feature extraction backbones (ResNet-18 and PointVector-S) and the MLP classifier are active during training. 
Input data flows through the backbones, features are extracted and concatenated, and then passed to the MLP. 
The crucial difference is that the loss is back-propagated through the entire network, allowing the weights of the ResNet-18 and PointVector-S backbones to be fine-tuned simultaneously with the training of the MLP layers. 
This allow the feature extractors to potentially adapt their representations for the specific fusion task.

\section{Experiments}
\label{sec:experiments}
To evaluate the effectiveness of our proposed methods for classifying forest biodiversity potential using remote sensing data, we conducted a series of experiments using the BioVista dataset. 
Our experiments aimed to compare the performance of 2D orthophoto-based classification, 3D ALS point cloud-based classification, and fusion approaches that combine both data modalities.
\added{
The primary focus of our analysis is the binary classification task distinguishing areas of high and low forest biodiversity potential.
For completeness, we also performed a three-class experiment incorporating a medium-potential category (HNV scores 4–6) to examine the increased difficulty of distinguishing among high, medium, and low biodiversity levels.
This additional experiment is presented only in this section, as we regard the binary setup as the principal and most meaningful configuration.
Accordingly, we generated \num{26834} medium-potential samples (train: \num{16080}, validation: \num{5380}, test: \num{5374}) using the procedure outlined in Section~\ref{sec:dateset_creation}.
}

\subsection{Evaluation metrics}
\label{subsec:evaluation-metrics}
We track the overall accuracy and the mean accuracy for assessing the performance of the various classifiers' ability to correctly identify high- and low forest biodiversity potential samples.

Overall accuracy (OAcc) is the most straightforward measure of a classifier's performance. 
It is defined as \added{$\text{OAcc} = N_\text{correct}/N_\text{total}$}, the ratio of correctly classified samples, $N_\text{correct}$, to the total number of samples, $N_\text{total}$.
While OAcc provides a general sense of the model's performance, it can be misleading in cases of class imbalance. 
Mean accuracy (MAcc) addresses the potential bias in overall accuracy due to class imbalance and is calculated as the average of the accuracies for each class, thereby, weighing the significance of each class equally important. 
\added{In the binary classification scenario}, we calculated the accuracy for classifying low forest biodiversity potential samples, $\text{Acc}_{\text{low}}$ and accuracy for classifying high forest biodiversity potential samples, $\text{Acc}_{\text{high}}$ and define the MAcc as follows:
\[
\text{MAcc} = \frac{\text{Acc}_{\text{high}} + \text{Acc}_{\text{low}}}{2}
\]

\subsection{2D classification performance evaluation}
\label{subsec:2d-classification-performance-evaluation}
For the classification of forest biodiversity potential using 2D orthophotos, we employed the ResNet-18 architecture~\citep{kaiming-resnet}. 
Deeper variants such as ResNet-34 and ResNet-50 were explored in preliminary experiments, but they did not yield significant performance improvements, leading us to select the more computationally efficient ResNet-18. 

We set the maximum training epochs to 20, as initial training experiments showed a validation performance plateau after approx. 10--20 epochs of training.
We used cross-entropy loss and employed an AdamW optimizer and a cosine annealing learning rate scheduler~\citep{2017-loshchilov-adamw,2016-loshchilov-cosine-annealing-lr}.

\begin{table}[t]
\footnotesize
\caption{Hyperparameter tuning of ResNet-18. 
Overview of the hyperparameters evaluated in the experiments, including the ranges explored and the final selected values.}
\label{tab:resnet-hyperparameters}
\centering
\begin{tabular}{@{}llc@{}}
\hline
\textbf{Hyperparameter} & \textbf{Range Tested} & \textbf{Final Value}  \\ \hline
Batch Size             & [4, 8, 16, 32, 64]                    & 8                     \\ \hline
Learning Rates         & [0.01, 0.001, 0.0001, 0.00001, 0.000001] & 0.0001               \\ \hline
Weighted Loss          & [True, False]                       & True                  \\ \hline
Pretrained             & [True, False]                       & False                  \\ \hline
\end{tabular}

\end{table}

We evaluated the combinations of hyperparameters listed in Table~\ref{tab:resnet-hyperparameters}, which also show the final selected values.
The missing benefits of using pre-trained weights from ImageNet might stem from the fact that the aerial perspective and characteristics of orthophotos differ substantially from the object-centric images of the ImageNet dataset. 

The input images have a resolution of 240×240 pixels, maintaining the original 12.5 cm ground sampling distance of the orthophotos.
As we employed the sample plot size and shape standards of the Danish NFI~\citep{alban-nfi-field-instructions}, pixels outside the 30-meter diameter circle are set to zero for all color channels.

To enhance model robustness and prevent over-fitting, we evaluated several data augmentation techniques during training, with inspiration from \citep{li2023treeheightprediction} and \citep{tolan2024canopy-height-maps}:
\begin{itemize}
    \item Random rotations from 0--360 degrees to account for rotational invariance in aerial imagery.
    \item Random color jittering with brightness, saturation, and hue adjustments to account for varying lighting conditions at capture times.
    \item Random Gaussian blur with sigma values (0.1--2.0) and kernel size of 5$\times$5 pixels, which could help improve model robustness to image quality differences.
\end{itemize}

We observed a significant performance increase when applying Gaussian blur during training. 
This could potentially be due to the presence of noise in the orthophotos which is reduced or removed by the Gaussian blur operation.  
We skipped random rotations and color jittering completely in the final experiments, as we did not observe a positive influence on the validation performance. 

When averaged over seven runs with the best identified hyperparameter settings and trained and evaluated on the RGB-color channels of the orthophotos, ResNet-18 achieved an overall accuracy of \replaced{$77.6\%$}{$75.5\%$} on the BioVista test set \replaced{on the binary classification task}{and Table~\ref{tab:results}}\comment{\small table updated}. 
The model performed notably better on the high forest biodiversity potential class (\replaced{$88.0\%$}{$86.4\%$}) compared to the low forest biodiversity potential class (\replaced{$67.5\%$}{$64.9\%$}), indicating a significant class imbalance in the predictive capabilities of the model, which we address in Section~\ref{sec:discussion}.

\added{
 As expected, this task proved more challenging: the confusion matrix computed over seven trials  in Table~\ref{tab:cm}a shows that while the high-potential class can still be identified reliably, the low-potential class is frequently confused with the medium-potential class.
}

\subsubsection{Near-infrared for forest biodiversity assessment}
\label{subsucsec:near-infrared-in-orthophotos}
In addition to conducting experiments with 2D orthophotos using the RGB color channels, we further evaluated ResNet-18 trained on orthophotos \added{using the RGB channels plus the near-infrared (NIR) channel. Including the NIR channel consistently improved performance.
On the binary task, ResNet-18 using RGB and NIR achieved an overall accuracy of $77.6\%$, see Table~\ref{tab:results}.}

\deleted{with various combination of color channels including the near-infrared channel, as show in Figure~\ref{fig:performance-comparison-color-channels}.
When comparing models trained on a only a single color channel, R, G, B or NIR, we see the best performance using the NIR-channel and the worst performance using the red channel.}
\deleted{We observed a clear pattern -- models trained on orthophotos with color channels which includes NIR outperforms models trained on color channels excluding the NIR channel.
All of the results seen in Figure~\ref{fig:performance-comparison-color-channels} are averages of at least 7 runs where each run is trained for 20 epochs, using the same hyperparameter settings as described in Section~\ref{subsec:2d-classification-performance-evaluation}.
We saw the best performance among the evaluated combinations to be models trained on NIR, green and blue channels (NGB), with an overall accuracy of $76.7\%$, as seen in Table~\ref{tab:results}.
For the remainder of this work, 2D ResNet-18 will refer to instances trained on NGB.}

\subsection{3D classification performance evaluation}
\label{subsec:3d-classification-performance-evaluation}
For the classification of forest biodiversity potential using 3D ALS point cloud, we employed the PointVector architecture. 
We used the x, y, z coordinates from the ALS point clouds.
Additionally, we added the relative height (h) as an input feature, similar to~\citep{2019-thomas-hugues-kpconv}, derived by subtracting all z coordinates of a point cloud with its min z-value. 

We trained PointVector for 20 epochs, using cross-entropy loss and the AdamW optimizer and cosine annealing learning rate scheduler, similar to the original PointVector paper~\cite{deng-pointvector}.

\begin{table}[th]
\caption{Hyperparameter tuning of PointVector-S. Overview of hyperparameters evaluated during the experiments with PointVector trained and tested on the BioVista dataset. The final column represents the final selected values.}
\footnotesize
\centering
\begin{tabular}{@{}llc@{}}
\hline
\textbf{Hyperparameter}   & \textbf{Range Tested} & \textbf{Final Value}  \\ \hline
Batch Size                & [4, 8, 12, 16]              & 8                \\ \hline
Learning Rates            & [0.001, 0.0001, 0.00001]    & 0.0001           \\ \hline
Number of Points          & [8.192, 16.384, 24.576]     & 16.384           \\ \hline
Query Ball Radius         & [0.6, 0.65, 0.7, 0.75, 0.8] & 0.65             \\ \hline
Query Ball Radius Scaling & [1.25, 1.5, 2.0]            & 1.5              \\ \hline 
Weighted Loss             & [True, False]               & False          \\ \hline
\end{tabular}

\label{tab:pointvector-hyperparameters}
\end{table}

We carefully evaluated combinations of the hyperparameter settings seen from Table~\ref{tab:pointvector-hyperparameters} which also displays the final selected values.
The batch size and weighted loss settings had a very minor impact on validation performance, while learning rate, query ball radius, and query ball radius scaling had a large impact on the validation performance, with overall accuracy on the validation set ranging from 71.6\% with the worst settings to 76.1\% with the best settings. 
For the query ball radius, we initially set it to 0.7 using the method by~\cite{jensen-data-driven-tuning-2023} and evaluated nearby values, resulting in the final selection of 0.65. 
The performance of experiments in which we sampled 24.576 points from each point cloud actually performed slightly better than for 16.384. 
However, we chose to go with sampling 16.384 points, as we deemed the trade-off between performance gains and the additional required computational cost not worthwhile. 

We evaluated the data augmentation techniques employed by the original PointVector and PointNeXt paper~\citep{guocheng-pointnext}, which include random rotation around the z-axis (up to 360 degrees), random scaling, and random point jittering. 
Surprisingly, we found that none of the augmentations had a positive influence on the performance, resulting in the fact that they were skipped for the final evaluation.

\added{On the binary task,} using the hyperparameter settings from Table~\ref{tab:pointvector-hyperparameters}, the PointVector-S model achieves an overall accuracy of $75.8\% \pm 0.4$ on the BioVista test dataset, when averaged over 7 runs, as seen in Figure~\ref{fig:barplot-high-and-low-correct-and-incorrect} and Table~\ref{tab:results}, which was the lowest performance achieved by the tested methods. 
\added{When distinguishing all three forest biodiversity potential classes, the performance dropped significantly, see
the confusion matrix in Table 7 b). Both low and high potential
areas are often confused with the medium potential class.}





\subsection{2D-3D classification performance evaluation}
\label{subsec:2d-3d-classification}
To classify forest biodiversity potential using both 2D orthophotos \deleted{(NGB channels)} and 3D ALS point clouds, we evaluated the three multimodal approaches outlined in Section~\ref{sec:method}: confidence-based ensembling, MLP fusion with \replaced{frozen backbones}{pre-computed features}, and end-to-end training.

\begin{figure}[t]
   \centering
   \includegraphics[width=.49\linewidth]{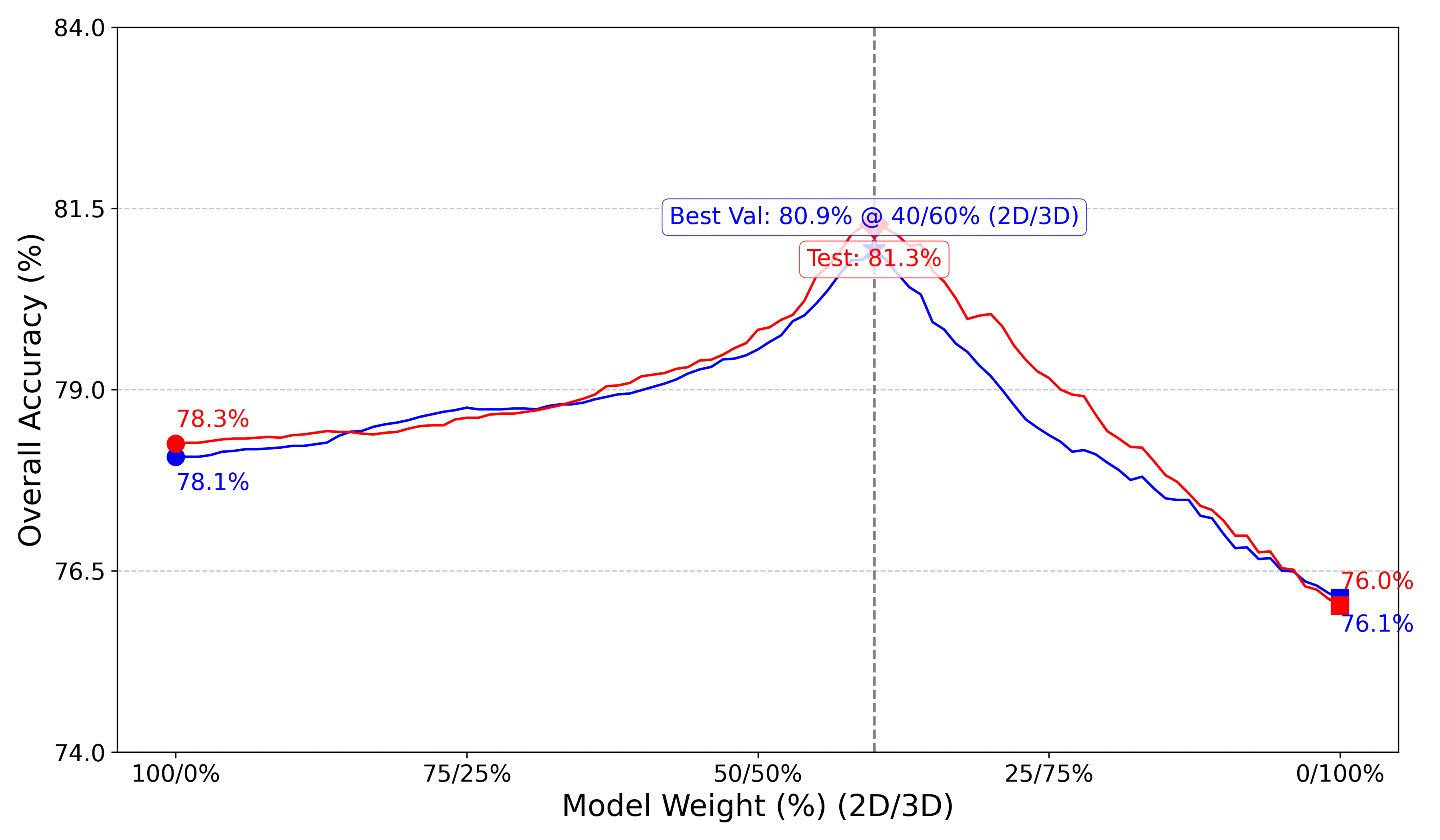}\hfill%
   \includegraphics[width=.49\linewidth]{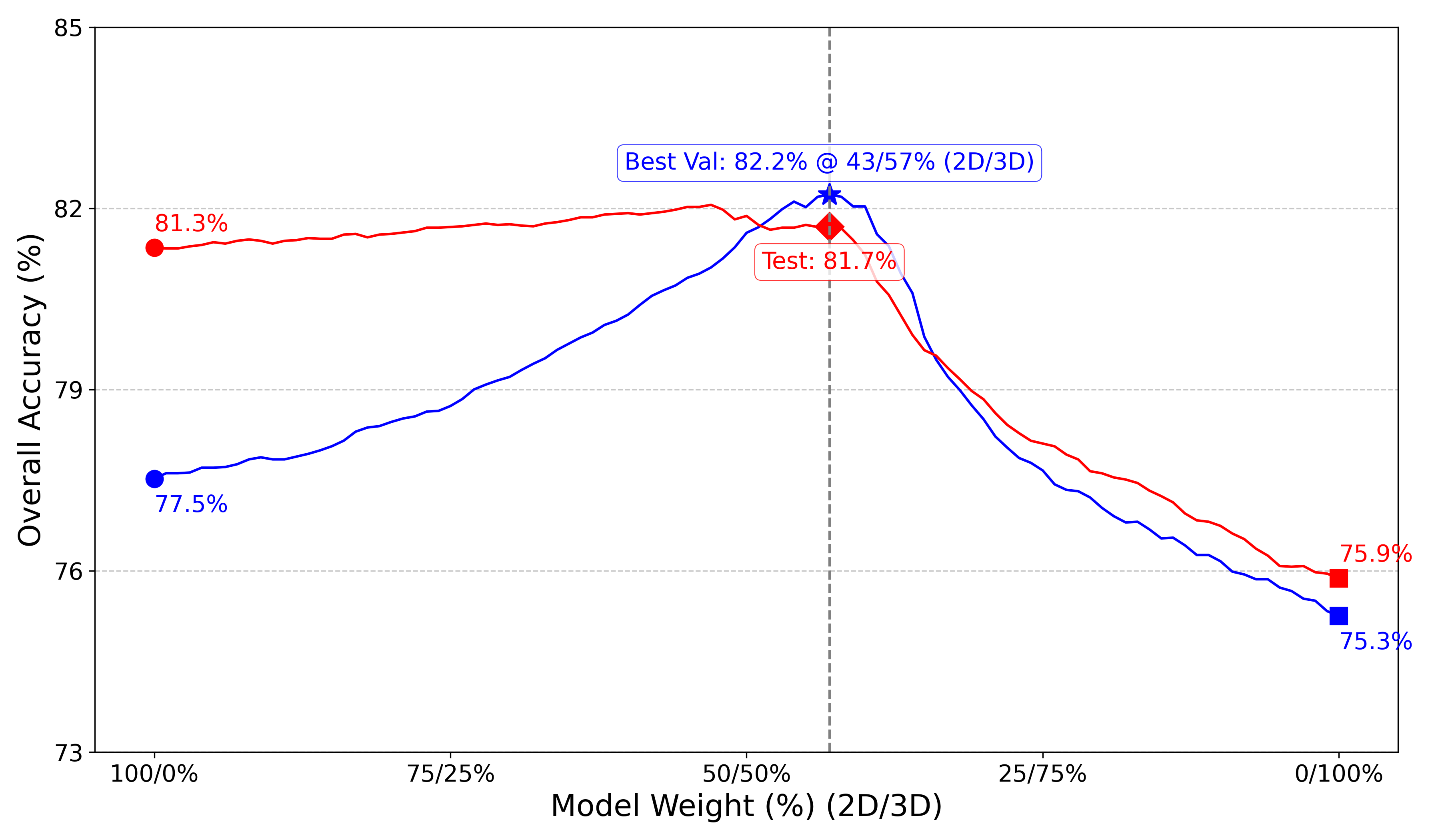}
   \caption{Confidence-based ensembling.
   Performance for two distinct pairs of trained 2D (ResNet-18) and 3D (PointVector-S) models. \textbf{(a)} Results for the first pair. \textbf{(b)} Results for the second pair. Each plot shows the overall classification accuracy on the BioVista validation set (blue curve) and test set (red curve) as the weighting shifts from 100\% ResNet (2D) / 0\% PointVector (3D) to 0\% 2D / 100\% 3D. 
   The optimal weight (vertical dashed line) is selected based on the peak validation accuracy (blue star), and the resulting test set accuracy at this optimal weight is indicated (red diamond).}
   \label{fig:ensempling-validation-and-test-performance}
\end{figure}

\textbf{Confidence-based ensembling:} For the confidence-based ensembling approach (Figure~\ref{fig:method-overview} (c)), we combined the confidence outputs from seven distinct instances of trained ResNet-18 models and seven distinct instances of trained PointVector-S models. 
We used weighted averaging, where the weights ($w_{\text{2D}}$ and $w_{\text{3D}}$) were optimized based on performance on the BioVista validation set, as illustrated in Figure~\ref{fig:ensempling-validation-and-test-performance}. 
Generally, this optimization resulted in a slight weighting preference towards the 3D PointVector predictions, with typical weight ratios ranging from 45:55 to 40:60 (ResNet:PointVector).
The weight ratio is surprising, as PointVector evaluated on 3D ALS point clouds achieved worse performance than ResNet evaluated on 2D orthophotos. 
As this method leverages the outputs of already trained models, no additional training phase was required beyond determining the optimal weights. 
This ensemble approach achieved an overall accuracy of \replaced{$79.9\% \pm 1.1$}{$80.5\% \pm 2.0$} and a mean accuracy of \replaced{$80.1\% \pm 1.1$}{$80.6\% \pm 1.9$} on the test set. 
\deleted{It demonstrated the strongest performance on the high forest biodiversity potential class, achieving $89.5\% \pm 2.7$ accuracy (Table~\ref{tab:results}).}

\textbf{MLP Fusion (\replaced{frozen backbones}{pre-computed}):} 
For the MLP Fusion approach using \replaced{frozen backbones}{pre-computed features} (Figure~\ref{fig:method-overview}(d)), \replaced{input samples are passed through the non-trainable ResNet-18 and PointVector-S backbones during training and evaluation to generate 512-dimensional feature embeddings, which are then fed into the MLP classifier. Only the MLP parameters are updated while the backbone weights remain fixed. We performed extensive hyperparameter tuning for the MLP using these dynamically generated features.}{we first extracted 512-dimensional feature vectors from the frozen backbones of the trained ResNet-18 and PointVector-S models for all samples in the training, validation, and test sets. We then conducted extensive hyperparameter tuning for the MLP classifier using these static features.} 
The evaluated hyperparameters, their tested ranges, and the final selected values that yielded the best performance on the validation set are summarized in Table~\ref{tab:mlp-fusion-hyperparameters}.

\begin{table}[th]
\footnotesize
\centering
\caption{Hyperparameter tuning of the MLP Fusion (\replaced{frozen backbones}{pre-computed}) classifier. Overview of the hyperparameters evaluated, the ranges explored, and the final selected values yielding the best validation performance.}
\label{tab:mlp-fusion-hyperparameters}
\begin{tabular}{@{}llc@{}}
\hline
\textbf{Hyperparameter} & \textbf{Range Tested} & \textbf{Final Value} \\ \hline
Batch Size              & [8, 32, 64, 128, 256]                                  & 64             \\ \hline
Learning Rate           & [$10^{-2}$, $10^{-4}$, $10^{-6}$, $10^{-8}$]             & $10^{-4}$        \\ \hline
Weighted Loss       & [True, False]                                          & True           \\ \hline
MLP Hidden Layers       & (512); (512, 256); (1024, 512, 256)                     & (1024, 512, 256) \\ \hline
Dropout Rate            & [0.0, 0.3, 0.5]                                        & 0.0            \\ \hline
\end{tabular}
\end{table}

\begin{figure}[t]
   \centering
   \includegraphics[width=1\textwidth]{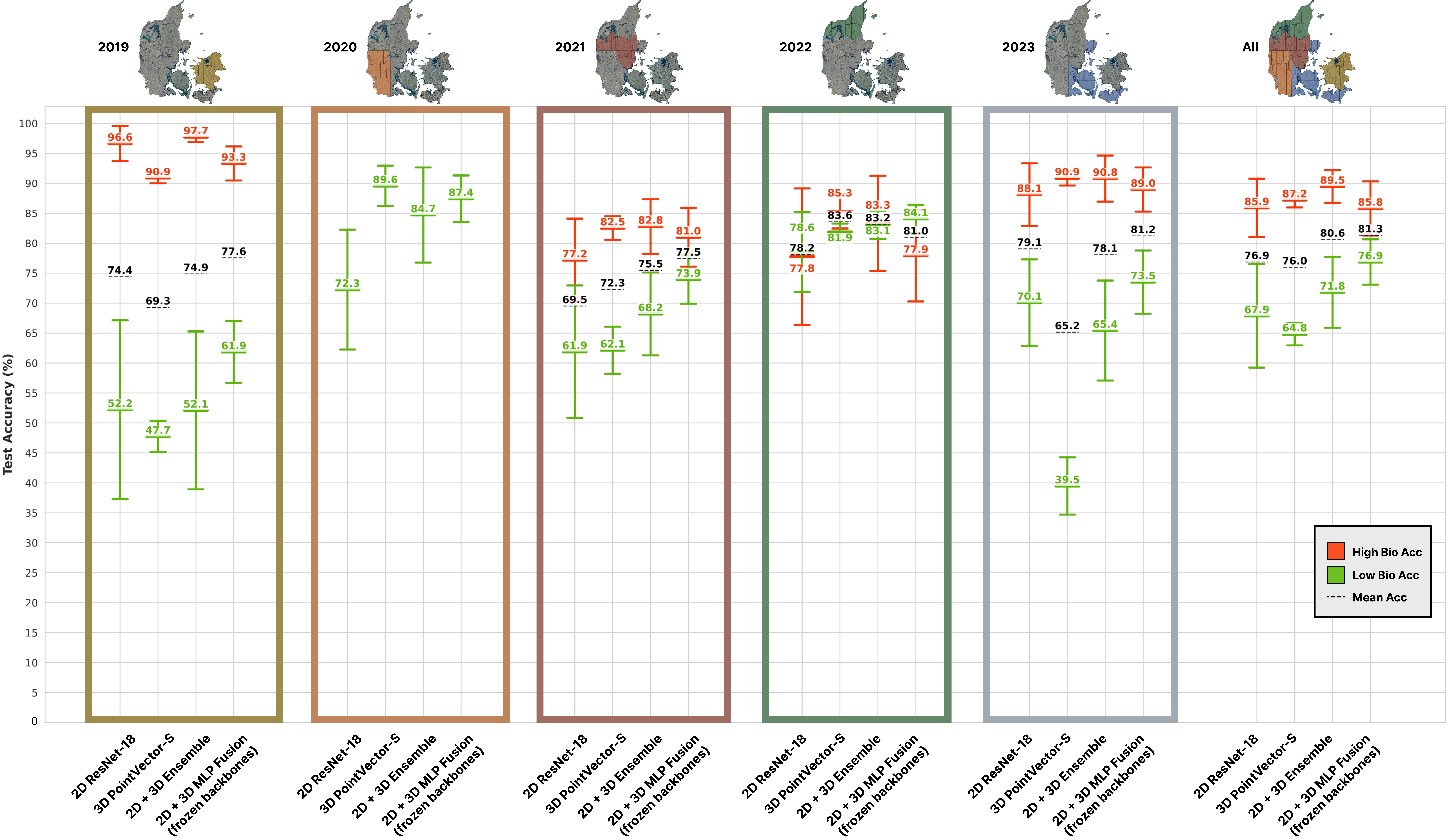}
   \caption{Regional model performance comparison.
   The mean test accuracy and accuracy per class across different regions in Denmark, for ResNet-18 trained and evaluated on 2D orthophotos, PointVector-S trained and evaluated on 3D ALS point clouds and two 2D and 3D multimodal fusion approaches.
   We train 7 instances of each model on the BioVista training set which spans the entire Denmark and the performance is an average of the 7 instances with accompanying standard deviations.}
   \label{fig:barplot-high-and-low-correct-and-incorrect}
\end{figure}

Training with the optimized settings detailed in Table~\ref{tab:mlp-fusion-hyperparameters} was conducted for 20 epochs using the AdamW optimizer. 
The \replaced{frozen backbones}{pre-computed} MLP fusion model yielded the best overall performance among all evaluated methods, achieving an overall accuracy of \replaced{$81.5\% \pm 1.0$}{$81.4\% \pm 0.5$} and a mean accuracy of \replaced{$81.7\% \pm 0.9$}{$81.3\% \pm 0.5$}. Notably, this approach produced a more balanced performance between classes compared to the single-modality models and the ensemble method, with \replaced{$91.3\% \pm 1.0$}{$85.8\% \pm 4.5$} accuracy on the high potential class and \replaced{$72.1\% \pm 2.6$}{$76.9\% \pm 3.8$} accuracy on the low potential class, as detailed in Table~\ref{tab:results} and visualized regionally in Figure~\ref{fig:barplot-high-and-low-correct-and-incorrect}.

\textbf{MLP Fusion (end-to-end):} Finally, we evaluated the end-to-end MLP Fusion approach (Figure~\ref{fig:method-overview} (e)), where both the MLP classifier and the ResNet-18 and PointVector-S backbones were trained simultaneously. 
For the MLP classifier part, we utilized the same optimal hyperparameters identified for the \replaced{frozen backbones}{pre-computed} approach (Table~\ref{tab:mlp-fusion-hyperparameters}). 
The backbones themselves were fine-tuned during this process. 
We experimented with different learning rates for the backbones, testing $10^{-4}$ (matching the MLP head), $10^{-5}$, and $10^{-6}$, finding that a lower learning rate of $10^{-5}$ for the backbones yielded the best results. 
Due to memory constraints associated with simultaneous training of all components, we employed gradient accumulation.
Gradients were accumulated over 8 steps using a base batch size of 8 to achieve an effective batch size of 64. 
As shown in Table~\ref{tab:results}, the end-to-end model achieved an overall accuracy of \replaced{$82.0\% \pm 0.7$}{$80.4\% \pm 0.2$} and mean accuracy of \replaced{$82.2\% \pm 0.7$}{$80.5\% \pm 0.2$}. 
\replaced{The end-to-end model performed overall best and showed the low standard deviation. The low forest biodiversity potential is $91.1\% \pm 0.7$ and high forest biodiversity potential of $73.2\% \pm 1.9$ — nearly matching the frozen-backbone model in the latter case. Hence, it led in all categories except the high-forest score, where results were similar.}{While its overall performance was slightly below the pre-computed MLP fusion, it achieved the highest accuracy on the low forest biodiversity potential class among all methods ($77.3\% \pm 2.1$). 
Furthermore, the very low standard deviations across the seven runs (e.g., $\pm 0.2\%$ for OAcc and MAcc) indicate that this end-to-end training regime resulted in the most stable and consistent model performance.}
The performance of the end-to-end trained MLP Fusion model (Figure~\ref{fig:method-overview} (e)) is also included in Table~\ref{tab:results} for comparison and regionally in Figure~\ref{fig:barplot-high-and-low-correct-and-incorrect}.
    
\begin{table}[t]
\footnotesize
\centering
\caption{\added{Overall model performance on the binary classification task. The performance metrics of our classification models evaluated on the BioVista test set, including overall accuracy (OAcc), mean accuracy (MAcc), accuracy for the high forest biodiversity potential class ($\text{Acc}_{\text{high}}$), and accuracy for the low forest biodiversity potential class ($\text{Acc}_{\text{low}}$). All metrics are reported as means with associated standard deviation across seven \added{model} runs with the same hyperparameter settings, {which are all} trained on the entire BioVista training set.}}
\label{tab:results}
{
\begin{tabular}{@{}lrrrr@{}}
\toprule
\textbf{Model} & \multicolumn{1}{c}{\textbf{OAcc}} & \multicolumn{1}{c}{\textbf{MAcc}} & \multicolumn{1}{c}{$\textbf{Acc}_{\text{high}}$} & \multicolumn{1}{c}{$\textbf{Acc}_{\text{low}}$} \\\midrule
2D ResNet-18  & 77.6\% ± 1.7 & 77.8\% ± 1.7 & 88.0\% ± 2.5 & 67.5\% ± 3.7 \\
3D PointVector-S & 75.8\% ± 0.4  & 76.0\% ± 0.4 &  87.2\% ± 1.2 &  64.8\% ± 1.9 \\
\addlinespace
2D  + 3D Ensembling & 79.9\% ±  1.1 & 80.1 \% ±  1.1 & 89.6\% ±  2.0 & 70.6\% ±  2.1 \\
2D  + 3D MLP Fusion (frozen backbones) & 81.5\% ± 1.0 & 81.7\% ± 0.9 & \textbf{91.3}\% ± 1.0 & 72.1\% ± 2.6  \\
2D  + 3D MLP Fusion (end-to-end) & \textbf{82.0\%} ± 0.7 & \textbf{82.2\%} ± 0.7 & {91.1\%} ± 0.7 & \textbf{73.2}\% ± 1.9 \\ 
\bottomrule
\end{tabular}
}
\end{table}

{\color{blue}
\begin{table}[h]
    \centering
    \caption{\added{Row-normalized confusion matrices of a) ResNet-18,  b) PointVector, c) PointVector and ResNet-18 combined with frozen backbones, and d) PointVector and ResNet-18 combined trained end-to-end for three classes and averaged over seven trials.\label{tab:cm}}}

\begin{minipage}{.49\textwidth}
    \begin{tabular}{crccc}
      & & \multicolumn{3}{c}{predicted}\\
            &   & {high} & {low} & {med.} \\
\multirow{3}{*}[-0.4ex]{\rotatebox{90}{actual}} & high   & \cellcolorval{71.3} 71.3 & \cellcolorval{7.7} 7.7 & \cellcolorval{21.0} 21.0 \\
& low    & \cellcolorval{22.9} 22.9 & \cellcolorval{32.2} 32.2 & \cellcolorval{45.0} 45.0 \\
& medium & \cellcolorval{23.4} 23.4 & \cellcolorval{13.6} 13.6 & \cellcolorval{63.0} 63.0 \\
     \multicolumn{5}{c}{a) ResNet-18}\\
    \end{tabular}
\end{minipage} 
\begin{minipage}{.49\textwidth}
    \begin{tabular}{crccc}
      & & \multicolumn{3}{c}{predicted}\\
      &   & {high} & {low} & {med.} \\
      \multirow{3}{*}[-0.4ex]{\rotatebox{90}{actual}} &
        high   & \cellcolorval{53.1} 53.1 & \cellcolorval{5.7} 5.7 & \cellcolorval{41.2} 41.2 \\
       & low    & \cellcolorval{21.1} 21.1 & \cellcolorval{30.6} 30.6 &  \cellcolorval{48.3} 48.3 \\
     &   medium & \cellcolorval{36.6} 36.6 & \cellcolorval{13.6} 13.6 & \cellcolorval{49.8} 49.8 \\
     \multicolumn{5}{c}{b) PointVector}\\
    \end{tabular}
\end{minipage} 

\begin{minipage}{.49\textwidth}
    \begin{tabular}{crccc}
      & & \multicolumn{3}{c}{predicted}\\
      &   & {high} & {low} & {med.} \\
\multirow{3}{*}[-0.4ex]{\rotatebox{90}{actual}} & high   & \cellcolorval{73.1} 73.1 & \cellcolorval{3.7} 3.7 & \cellcolorval{23.2} 23.2 \\
& low    & \cellcolorval{15.5} 15.5 & \cellcolorval{38.6} 38.6 & \cellcolorval{45.9} 45.9 \\
& medium & \cellcolorval{19.0} 19.0 & \cellcolorval{11.5} 11.5 & \cellcolorval{69.5} 69.5 \\
     \multicolumn{5}{c}{c) Fusion (frozen)}\\
    \end{tabular}
\end{minipage} 
\begin{minipage}{.49\textwidth}
    \begin{tabular}{crccc}
      & & \multicolumn{3}{c}{predicted}\\
      &   & {high} & {low} & {med.} \\
      \multirow{3}{*}[-0.4ex]{\rotatebox{90}{actual}} & high   & \cellcolorval{71.1} 71.1 & \cellcolorval{4.7} 4.7 & \cellcolorval{24.2} 24.2 \\
& low    & \cellcolorval{18.0} 18.0 & \cellcolorval{38.4} 38.4 & \cellcolorval{43.6} 43.6 \\
& medium & \cellcolorval{19.7} 19.7 & \cellcolorval{12.4} 12.4 & \cellcolorval{67.8} 67.8 \\
     \multicolumn{5}{c}{d) Fusion (end-to-end)}\\
    \end{tabular}
\end{minipage} 

\end{table}
}

\subsection{2D-3D multimodal \deleted{fusion} benefits}
\label{2d-3d-multimodal-fusion-benefits}
From Table~\ref{tab:results}, we see that the 2D + 3D \deleted{MLP fusion} model significantly outperformed the single modality 2D ResNet and 3D PointVector models \added{when considering the binary classification task}.
\added{As can be seen in Table~\ref{tab:cm}, the combined models also performed better in the three-category setting.}

The question arises whether the performance gain was not merely due to ensembling two models (e.g., ResNet-18 and PointVector). 
We verified that the improvement stems indeed from the fact that the 2D orthophoto and 3D ALS point cloud data modalities carry distinct and complementing features.
\added{Concretely,} we did this by comparing the performance of \replaced{ensembling}{MLP} models\deleted{, with settings as described in Section~\ref{subsec:2d-3d-classification}, trained on outputs from multiple combinations} of two ResNet models (2D + 2D) \replaced{,}{and} two PointVector models (3D + 3D) to the performance of the 2D + 3D \deleted{MLP fusion} model.
We used three distinct instances of ResNet models and three distinct instances of PointVector models, \added{all} trained with \added{exactly the same} hyperparameter settings, \replaced{which are shown in Table~\ref{tab:resnet-hyperparameters} and Table~\ref{tab:pointvector-hyperparameters}}{as described in Section~\ref{subsec:2d-classification-performance-evaluation} and~\ref{subsec:3d-classification-performance-evaluation}} respectively, which we refer to ResNet 1, ResNet 2 and ResNet 3 and PointVector 1, PointVector 2 and PointVector 3 in Table~\ref{tab:modality-fusion-comparison-results}\comment{updated table}.
\added{The only difference between the model instances are the random seeds which the model weights are initialized with when the model training starts}.
The ResNet models were trained on orthophotos using the \replaced{RGB and NIR}{NIR, G and B color} channels.
Then we compared the performance of standalone models against \replaced{ensembled}{fused} models. 
We calculated the percentage point gain of the \replaced{ensembled}{fused} model version compared to the average performance of the standalone models and to the best performing standalone model, referred to as gain (avg) and gain (max) in Table~\ref{tab:modality-fusion-comparison-results}. 

\begin{table}
\caption{\added{Multimodal \replaced{ensemble}{fusion} performance comparison. Analysis of model \deleted{fusion} performance between different modality pairs (2D-2D, 3D-3D, and 2D-3D), showing overall accuracy for individual models and fused versions along with performance gains relative to average and maximum single model accuracies.}}
\label{tab:modality-fusion-comparison-results}
\centering
\footnotesize  
\renewcommand{\arraystretch}{1.2}
\begin{tabular}{@{}llccccc@{}}
\toprule
\textbf{Model A} & \textbf{Model B} & \textbf{OAcc (A)} & \textbf{OAcc (B)} & \textbf{OAcc (Fused)} & \textbf{Gain (Avg)} & \textbf{Gain (Max)} \\
\midrule
\multicolumn{7}{c}{\textbf{2D-2D}}\\
\midrule
ResNet 1 & ResNet 2 & 78.3 \% & 77.8 \% & 79.0 \% & +1.0 & +0.8 \\
ResNet 2 & ResNet 3 & 77.8 \% & 81.3 \% & 80.5 \% & +0.9 & -0.8 \\
ResNet 3 & ResNet 1 & 81.3 \% & 78.3 \% & 80.9 \% & +1.1 & -0.5 \\
\cmidrule(l){3-7}
\multicolumn{2}{@{}l}{{Average}} & 79.1 \% & 79.1 \% & 80.1 \% & +1.0 & +1.0 \\
\midrule
\multicolumn{7}{c}{\textbf{3D-3D}}\\
\midrule
PointVector 1 & PointVector 2 & 76.0 \% & 76.0 \% & 76.1 \% & +0.1 & +0.1 \\
PointVector 2 & PointVector 3 & 76.0 \% & 75.9 \% & 76.0 \% & +0.0 & -0.0 \\
PointVector 3 & PointVector 1 & 75.9 \% & 76.0 \% & 76.4 \% & +0.5 & +0.4 \\
\cmidrule(l){3-7}
\multicolumn{2}{@{}l}{Average}  & 76.0 \% & 76.0 \% & 76.2 \% & +0.2 & +0.2 \\
\midrule
\multicolumn{7}{c}{\textbf{2D-3D}}\\
\midrule
ResNet 1 & PointVector 1 & 78.3 \% & 76.0 \% & 81.3 \% & +4.1 & +3.0 \\
ResNet 2 & PointVector 2 & 77.8 \% & 76.0 \% & 80.2 \% & +3.3 & +2.4 \\
ResNet 3 & PointVector 3 & 81.3 \% & 75.9 \% & 81.7 \% & +3.1 & +0.3 \\

\cmidrule(l){3-7}
\multicolumn{2}{l}{@{}Average} & 79.1 \% & 76.0 \% & 81.1 \% & +3.5 & +1.9 \\
\bottomrule
\end{tabular}

\end{table}

We observed that \replaced{ensembling models}{fusion of feature embeddings} from 2D orthophotos and 3D ALS point clouds leads to consistent performance gains, with a mean gain (avg) of \replaced{3.5}{4.4} percentage point improvement.
3D + 3D \deleted{fusion} does not show \added{strong} synergy, indicating that different instances of the PointVector model extracts consistent feature embeddings. 
2D + 2D \deleted{fusion} performs marginally better than individual models, indicating that different instances of trained ResNet models actually capture complementary features, while not as much as observed for 2D + 3D\deleted{ MLP fusion}. 

\begin{figure}[t]
   \centering
   \includegraphics[width=1.0\textwidth]{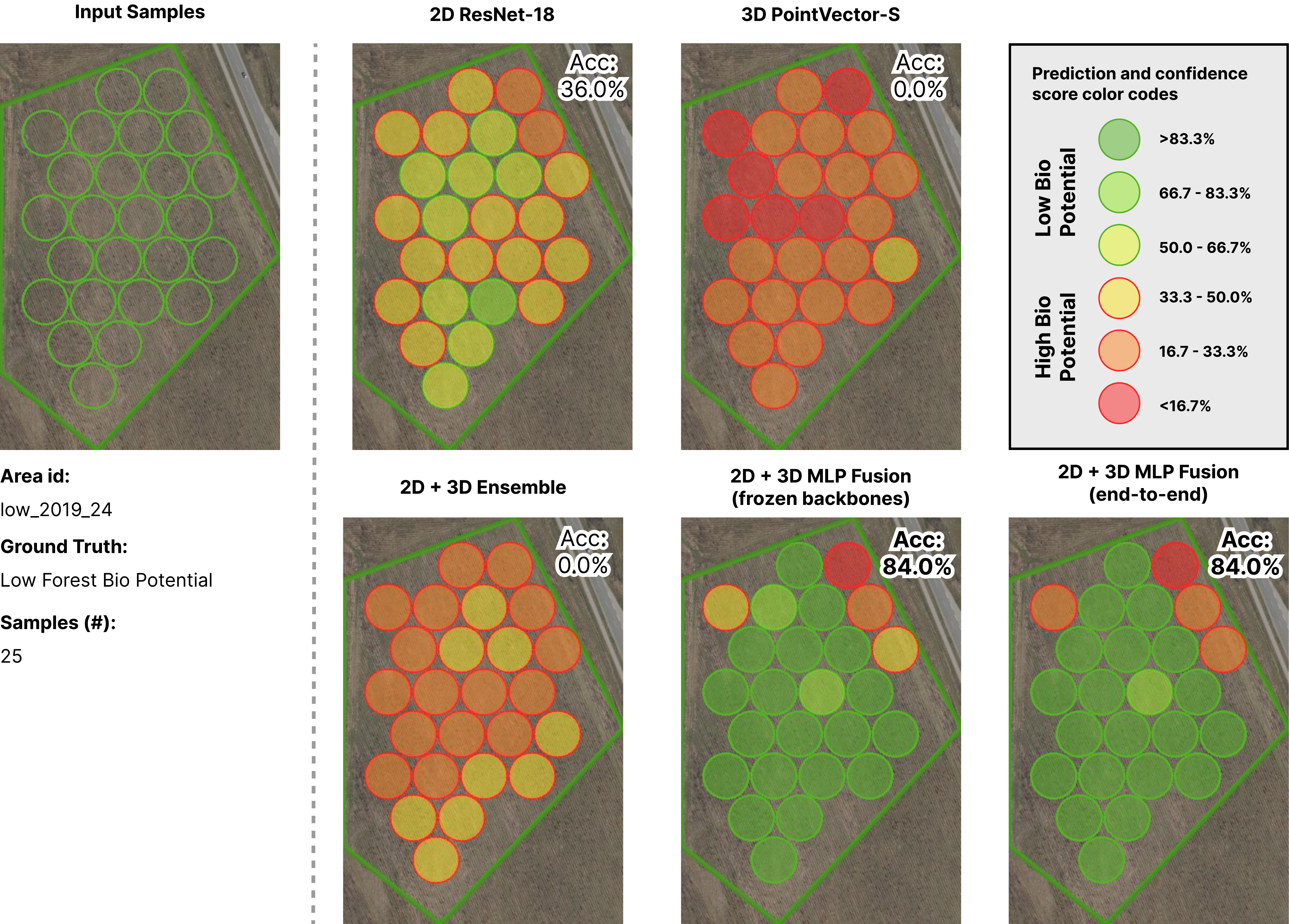}
   \caption{\added{Qualitative analysis --} low forest biodiversity potential example.
   Predictions and as associated confidence scores from from all tested methods for an area with low forest biodiversity potential (low\_2019\_24). The 2D + 3D MLP Fusion methods achieve the best performance with a mean accuracy of 84.0\%.
   Accuracies for each model are shown in the top right of each prediction image.} 
   \label{fig:qualitative-analysis-low-bio-example}
\end{figure}

\subsection{Qualitative analysis}
To better understand the tested methods' performance and behavior, we analyzed their predictions on two representative forest areas with contrasting biodiversity potential in Figures~\ref{fig:qualitative-analysis-low-bio-example} and \ref{fig:qualitative-analysis-high-bio-example}.
Figure~\ref{fig:qualitative-analysis-low-bio-example} shows predictions on a low forest biodiversity potential area (id: low\_2019\_24) from the BioVista test set, containing 25 samples. 
Both MLP Fusion methods (end-to-end trained and trained \replaced{with frozen backbones}{on pre-computed feature embeddings}) notably outperformed single-modality methods, achieving an accuracy of 84.0\% compared to ResNet's 36.0\% and PointVector's 0.0\%. 
The example highlights instances where MLP Fusion methods manage to correctly classify the target class, even when both single-modality methods fail to do so. 
Additionally, we observed that the end-to-end and \replaced{frozen backbone}{pre-computed} variants of MLP fusion generally yielded similar predictions with minor confidence score variations.

\begin{figure}[!t]
   \centering
   \includegraphics[width=0.85\textwidth]{images/figure-12-qualitative-analysis-high-bio-example.pdf}
   \caption{\added{Qualitative analysis --} high forest biodiversity potential example.
   Predictions and as associated confidence scores from all tested methods on a high forest biodiversity potential area (high\_2021\_24) from the BioVista test set\deleted{, except 2D + 3D MLP Fusion (end-to-end)}. 
   Accuracies for each model are shown in the top right of each prediction image.}
   \label{fig:qualitative-analysis-high-bio-example}
\end{figure}

Figure~\ref{fig:qualitative-analysis-high-bio-example} shows predictions on a high forest biodiversity potential area (id: high\_2021\_24) containing 216 samples. 
This example highlights that the ResNet model (using spectral data from 2D orthophotos) and the PointVector model (using structural data from 3D ALS point clouds) can produce significantly different performance for the same area.
The ResNet model achieves an accuracy of 74.1\% and PointVector 31.5\%.
Crucially, the MLP fusion methods, which integrates both data modalities, demonstrate the ability to leverage the more informative signal, in this case provided by the 2D orthophotos.

\begin{figure}[!t]
   \centering
   \includegraphics[width=1.0\textwidth]{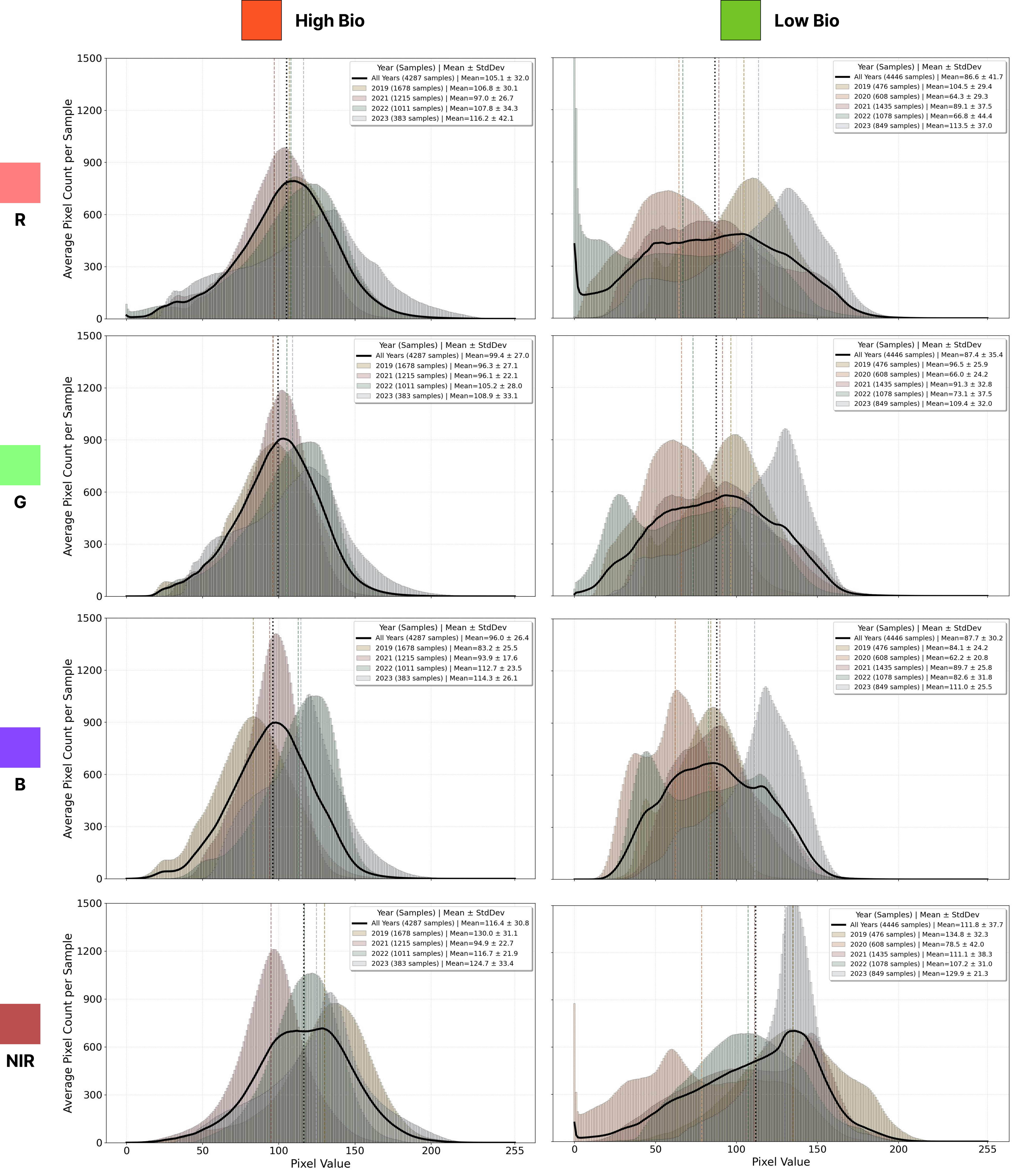}
   \caption{Pixel value distribution per class across regions.
   Comparison of the pixel value distributions for the high- and low forest biodiversity potential class of the red, green, blue, and near-infrared channels of the orthophotos from region to region in samples from the BioVista test dataset.} 
   \label{fig:color-channel-distribution-per-region-per-class}
\end{figure}

\section{Discussion}
\label{sec:discussion}
Our work demonstrates that remote sensing combined with deep learning offers a promising pathway for large-scale forest biodiversity \added{potential} monitoring, though important limitations and considerations must be acknowledged.

\subsection{Implications}
\added{A central conclusion of our work is that 2D- and 3D multimodal fusion provides a significant performance advantage against single modality methods. 
Our results clearly show that models fusing 2D orthophotos and 3D ALS point clouds consistently outperform models based on a single modality (see Table~\ref{tab:results}). 
Our analysis in Section~\ref{2d-3d-multimodal-fusion-benefits} further demonstrates that this is not merely an ensembling effect -- the performance gain stems from the fact that the spectral information from orthophotos and the structural information from ALS point clouds provide distinct and complementary features for assessing biodiversity potential.
Ensembling two models of the same type (e.g., 2D + 2D) yields only marginal gains compared to the substantial improvements seen with 2D + 3D fusion (see Table~\ref{tab:modality-fusion-comparison-results}).
}

\replaced{Our results, shown in Table~\ref{tab:results} and Figure~\ref{fig:barplot-high-and-low-correct-and-incorrect}}{Furthermore, our results}, demonstrate that deep learning models can effectively assess forest biodiversity potential on a coarse-grained level using remote sensing data, with our best model achieving an overall accuracy of \replaced{82.0\% when separating low and high biodiversity potential areas}{81.4\%}. 
\deleted{Particularly noteworthy is how combining 2D orthophotos and 3D ALS point clouds, through deep learning-based multimodal fusion, enhances classification performance.
This performance enhancement shows that spectral and structural information complement each other in meaningful ways for forest biodiversity assessment, as clearly demonstrated in Section~\ref{2d-3d-multimodal-fusion-benefits}.}

\added{
Our results in Table~\ref{tab:cm} justify the binarization of biodiversity potential, as experiments with three classes showed reduced performance and higher confusion between intermediate levels. This outcome suggests that the current combination of remote sensing data and the available biodiversity potential definitions may not yet be sufficient to capture finer gradients reliably.
Despite this simplification, binary predictions of biodiversity potential remain valuable for identifying and prioritizing areas for field surveys, conservation planning, and targeted data collection.
While more detailed and harmonized biodiversity potential metrics would be desirable, such scalable indicators are not yet available across Europe or Scandinavia. The current HNV framework, though imperfect, serves as a practical and reproducible proxy for large-scale biodiversity assessments and provides a solid foundation for future refinement as improved data and indicators emerge.
}

\subsection{Limitations}
Several important limitations of our approach warrant discussion. 
Firstly, the use of the HNV forest map as ground truth presents both strengths and limitations. 
While the map relies on proxy features rather than direct biodiversity measurements, it has undergone rigorous expert validation and systematic testing that supports its utility for identifying forest with high biodiversity potential. 

Another limitation pertains to the seasonality of the orthophoto and ALS point cloud data. 
As noted in Section~\ref{subsec:2d-orthophotos}, the images were acquired between March 1st and May 1st, corresponding to the leaf-off period for many deciduous species in Danish temperate forests. 
We acknowledge that forest canopy structure and spectral responses vary significantly across seasons. 
Utilizing orthophotos and ALS point clouds captured during the summer or autumn would provide different spectral information related to chlorophyll content, vegetation health, and potentially allow for better or worse differentiation based on leaf characteristics.
This seasonal variation represents a form of temporal or concept drift, which is a known challenge for machine learning models operating on environmental data over time~\citep{nikolov-2021-seasons}.

We observed interesting patterns in regional performance variations, particularly for the low forest biodiversity potential class, for which our methods exhibited significantly worse performance across all regions compared to the high forest biodiversity potential class, as seen in Figure~\ref{fig:barplot-high-and-low-correct-and-incorrect}, except for the 2022 region. 
One reason for this variance likely stems from the heterogeneous nature of low forest biodiversity samples, which can exhibit various combinations of HNV proxy features, while high forest biodiversity samples share presence of many or all of the same proxy features. 
We illustrate this in Figure~\ref{fig:color-channel-distribution-per-region-per-class}, where it becomes very clear that the low forest biodiversity samples exhibits a large variation from region to region across all color channels of the orthophoto samples in the BioVista test dataset.  
In addition, for region 2019, in which we observe the worst performance on the low biodiversity class across all methods, the class imbalance is significant in favor of the high biodiversity class, as seen in Figure~\ref{fig:color-channel-distribution-per-region-per-class}. 
This suggests that the number of low biodiversity samples in the BioVista dataset would have to be increased, in order for the our methods to capture the greater distributional variation of the class.
The low performance of the PointVector-S model on the low biodiversity class in region 2023 suggests that variations in ALS point clouds also warrant further investigation.

\subsection{Future work}
A crucial next step is to evaluate the generalizability of our approach across different forest biomes. 
While our method shows promise for temperate forests in Denmark, tropical, boreal, and other forest types present distinct challenges and characteristics. 
Tropical forests, for instance, typically exhibit much higher species diversity and more complex canopy structures than temperate forests, which could affect both the ALS point cloud patterns and spectral signatures in orthophotos.
\added{Although the model is robust to terrain variation, incorporating absolute elevation as an additional input could be beneficial in more topographically complex regions, since it currently only learns from plot-relative height information.}
While our binary classification approach provides a \replaced{proof of concept}{solid foundation} for forest biodiversity \added{potential} assessment, future work should explore more fine-grained classification schemes. 
\deleted{Rather than categorizing forests as having high- or low biodiversity potential, a} A multi-class approach \replaced{that reliably distinguishes}{could distinguish} between several levels of biodiversity \deleted{potential. 
This} would provide more nuanced insights for conservation prioritization and forest management.

\section{Conclusion}
\label{sec:conclusion}
This study demonstrates the potential of 2D orthophotos, 3D ALS point clouds and a combination of the two, for assessment of forest biodiversity potential using deep learning methods.
Through evaluation using our introduced BioVista dataset, we have shown that deep learning methods can successfully classify forest biodiversity potential, with our best fusion model achieving an overall accuracy of \replaced{82.0\% when separating low and high potential areas}{81.4\%}. 
The superior performance achieved through multimodal fusion suggests that combining structural and spectral information provides a more comprehensive and reliable assessment of forest biodiversity.
This multimodal potential should be tested for predicting other forest-related variables as well. 
This research represents a step towards developing methods that can support forest biodiversity monitoring and conservation efforts, but further improvements in accuracy, \added{granularity}, and robustness are needed before these methods could be considered for practical applications. 
Future work should incorporate a more fine-grained classification scheme and test the methodology across different forest biomes.

\section{Acknowledgements and funding}
\label{sec:acknowledgements}
This research was enabled by support and funding provided by Ambolt AI, AI Denmark and The Pioneer Centre for Artificial Intelligence (DNRF grant number P1). 
This work has been partially supported by the Spanish project PID2022-136436NB-I00 and by ICREA under the ICREA Academia programme. 
CI acknowledges support from the Danish National Research Foundation (DNRF) through TreeSense, the Center for Remote Sensing and Deep Learning of Global Tree Resources (DNRF192).
The authors declare no conflict of interest.

\section{CRediT authorship contribution statement}
\textbf{Simon B.~Jensen:} Conceptualization, Data curation, Methodology, Writing – original draft, Writing – review and editing.
\textbf{Stefan Oehmcke:} Conceptualization, Writing – original draft, Writing – review and editing.
\textbf{Andreas Møgelmose:} Conceptualization, Writing – review and editing, Supervision.
\textbf{Meysam Madadi:} Conceptualization, Writing – review and editing, Supervision.
\textbf{Christian Igel:} Conceptualization, Writing – review and editing.
\textbf{Sergio Escalera:} Project administration, Writing – review and editing.
\textbf{Thomas B.~Moeslund:} Project administration, Funding acquisition, Writing – review and editing, Supervision.

\begin{appendices}

\section{Forest area performance}
We analyzed \added{binary} classification results on a selection of forest areas in the BioVista test set.
The results are shown in Table~\ref{tab:forest-area-results}.
The results reveal considerable variation in model performance across different areas. 
For some areas, all tested models display perfect classification accuracy, while others prove challenging for some or all methods.

For high forest biodiversity areas, we observed that several areas (e.g., high\_2019\_51 through high\_2019\_59) achieved 100\% accuracy across all models, suggesting these forest areas have clearly distinguishable characteristics of high forest biodiversity potential. 
However, some areas (e.g., high\_2022\_27 and high\_2022\_28) showed consistently poor performance across all models, indicating that certain high biodiversity features may be difficult to distinguish using current remote sensing approaches. 

We found similar patterns of variation in low forest biodiversity \added{potential} areas, where several areas (e.g., low\_2020\_31, low\_2023\_41) were classified perfectly by all models, while others (e.g., low\_2023\_29) proved particularly challenging with near-zero accuracy across all approaches. 
The MLP Fusion model showed notably strong performance on moderately difficult low forest biodiversity \added{potential} areas, often achieving the highest accuracy among all models.

\begin{table}[htbp] 
\tiny
  \centering
  \caption{Model performance on test areas.
Model performance comparison across forest areas in the BioVista test set. The table is divided into high- and low forest biodiversity potential areas, with each category showing the top 6, median 6 and bottom 6 forest areas sorted by average mean accuracy performance across all methods. 
For each forest area, the mean accuracy (MAcc) is shown for the tested models, with the best performing model(s) highlighted in bold.}
  \label{tab:forest-area-results}
  \resizebox{\textwidth}{!}{
  \begin{tabular}{@{}l r r r r r r r@{}}
    \toprule
    Forest Area Id & Samples & ResNet18 & PointVector-s & Ensembling & \multicolumn{1}{c}{MLP fusion} & \multicolumn{1}{c}{MLP fusion} & Model \\ 
    & & & & & \multicolumn{1}{c}{(\replaced{frozen backbones}{pre-computed})} & \multicolumn{1}{c}{(end-to-end)} & Avg. \\ 
    \midrule
       high\_2019\_51 & 8 & \textbf{100.0} & \textbf{100.0} & \textbf{100.0} & \textbf{100.0} & \textbf{100.0} & \textbf{100.0} \\
    high\_2019\_52 & 58 & \textbf{100.0} & \textbf{100.0} & \textbf{100.0} & \textbf{100.0} & \textbf{100.0} & \textbf{100.0} \\
    high\_2019\_53 & 41 & \textbf{100.0} & \textbf{100.0} & \textbf{100.0} & \textbf{100.0} & \textbf{100.0} & \textbf{100.0} \\
    high\_2019\_54 & 80 & \textbf{100.0} & \textbf{100.0} & \textbf{100.0} & \textbf{100.0} & \textbf{100.0} & \textbf{100.0} \\
    high\_2019\_56 & 76 & \textbf{100.0} & \textbf{100.0} & \textbf{100.0} & \textbf{100.0} & \textbf{100.0} & \textbf{100.0} \\
    high\_2019\_59 & 75 & \textbf{100.0} & \textbf{100.0} & \textbf{100.0} & \textbf{100.0} & \textbf{100.0} & \textbf{100.0} \\
    \midrule
    high\_2019\_61 & 80 & \textbf{100.0} & 81.3 & \textbf{100.0} & \textbf{100.0} & \textbf{100.0} & 96.9 \\
    high\_2019\_63 & 236 & 96.6 & 97.9 & \textbf{99.2} & 95.8 & 94.5 & 96.5 \\
    high\_2021\_28 & 448 & 95.5 & \textbf{100.0} & 99.1 & 96.0 & 92.2 & 95.9 \\
    high\_2021\_22 & 57 & \textbf{94.7} & \textbf{94.7} & 96.5 & 93.0 & 93.0 & 94.2 \\
    high\_2021\_25 & 55 & \textbf{94.6} & 87.3 & 92.7 & 92.7 & 90.9 & 91.5 \\
    high\_2019\_66 & 147 & 93.9 & \textbf{95.9} & \textbf{95.9} & 84.4 & 85.0 & 90.1 \\
    \midrule
    high\_2021\_24 & 216 & 74.1 & 31.5 & 63.0 & \textbf{74.5} & 63.0 & 61.3 \\
    high\_2022\_21 & 43 & \textbf{76.7} & 20.9 & 53.5 & 53.5 & 58.1 & 53.5 \\
    high\_2021\_27 & 29 & 31.0 & \textbf{48.3} & 34.5 & 31.0 & 31.0 & 34.5 \\
    high\_2021\_23 & 77 & 1.3 & \textbf{64.9} & 5.2 & 3.9 & 3.9 & 13.6 \\
    high\_2022\_27 & 18 & 0.0 & 0.0 & 0.0 & 0.0 & 0.0 & 0.0 \\
    high\_2022\_28 & 5 & 0.0 & 0.0 & 0.0 & 0.0 & 0.0 & 0.0 \\
    \midrule
    \midrule 
    low\_2020\_31 & 38 & \textbf{100.0} & \textbf{100.0} & \textbf{100.0} & \textbf{100.0} & \textbf{100.0} & \textbf{100.0} \\
    low\_2021\_54 & 45 & \textbf{100.0} & \textbf{100.0} & \textbf{100.0} & \textbf{100.0} & \textbf{100.0} & \textbf{100.0} \\
    low\_2021\_73 & 34 & \textbf{100.0} & \textbf{100.0} & \textbf{100.0} & \textbf{100.0} & \textbf{100.0} & \textbf{100.0} \\
    low\_2022\_41 & 69 & \textbf{100.0} & \textbf{100.0} & \textbf{100.0} & \textbf{100.0} & \textbf{100.0} & \textbf{100.0} \\
    low\_2023\_31 & 12 & \textbf{100.0} & \textbf{100.0} & \textbf{100.0} & \textbf{100.0} & \textbf{100.0} & \textbf{100.0} \\
    low\_2023\_41 & 23 & \textbf{100.0} & \textbf{100.0} & \textbf{100.0} & \textbf{100.0} & \textbf{100.0} & \textbf{100.0} \\
    \midrule
    low\_2023\_37 & 34 & 73.5 & \textbf{100.0} & \textbf{100.0} & 73.5 & 73.5 & 82.8 \\
    low\_2020\_33 & 33 & 69.7 & \textbf{93.9} & \textbf{93.9} & 78.8 & 78.8 & 82.8 \\
    low\_2021\_77 & 69 & 47.8 & \textbf{100.0} & \textbf{100.0} & 85.5 & 75.4 & 80.4 \\
    low\_2022\_42 & 49 & 77.6 & 75.5 & 79.6 & \textbf{81.6} & \textbf{81.6} & 79.9 \\
    low\_2023\_30 & 18 & 61.1 & \textbf{100.0} & 94.4 & 77.8 & 72.2 & 79.6 \\
    low\_2020\_29 & 49 & 77.6 & 75.5 & 77.6 & \textbf{81.6} & 79.6 & 78.9 \\
    \midrule
    low\_2023\_34 & 19 & 15.8 & 0.0 & 5.3 & 21.1 & \textbf{26.3} & 15.8 \\
    low\_2021\_64 & 90 & \textbf{18.9} & 2.2 & 11.1 & \textbf{18.9} & \textbf{18.9} & 14.8 \\
    low\_2019\_22 & 11 & 0.0 & \textbf{54.6} & 0.0 & 9.1 & 9.1 & 13.6 \\
    low\_2023\_29 & 7 & \textbf{14.3} & 0.0 & 0.0 & \textbf{14.3} & \textbf{14.3} & 9.5 \\
    low\_2023\_38 & 26 & 7.7 & \textbf{11.5} & 3.9 & 7.7 & \textbf{11.5} & 8.3 \\
    low\_2019\_27 & 1 & 0.0 & 0.0 & 0.0 & 0.0 & 0.0 & 0.0 \\
    \bottomrule
  \end{tabular}%
  } 
\end{table}

\section{Intermediate Fusion with Short Cuts}
\begin{figure}[ht]
    \centering
    \includegraphics[width=1.0\linewidth]{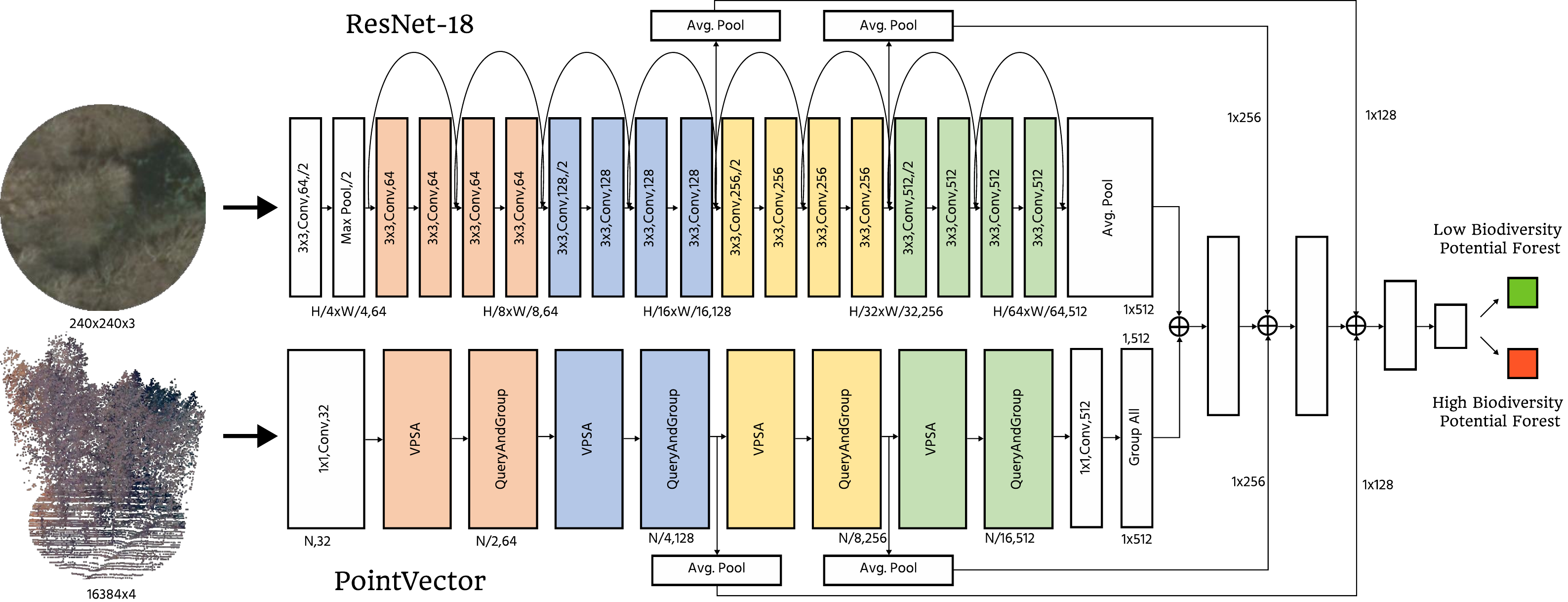}
    \caption{Intermediate fusion of 2D orthophotos and 3D ALS point clouds with short cuts. Feature embeddings from intermediate layers in the ResNet and PointVector backbones are pooled and concatenated into the fully connected layers of the MLP head, using short cuts.}
    \label{paper:g:fig:shortcut-fusion}
\end{figure}

In addition to the 2D and 3D fusion approaches presented in Section~\ref{sec:method} (late fusion through confidence-based ensembling and feature-level concatenation), we carried our an additional fusion experiments using intermediate fusion with short cuts. 
The approach is illustrated in Figure~\ref{paper:g:fig:shortcut-fusion}.

We train the network for 40 epochs, using the same hyperparameters as described in Table~\ref{tab:mlp-fusion-hyperparameters} and a learning rate of $10^{-5}$ for the ResNet and PointVector backbones, similar to our MLP Fusion (end-to-end). 
In order to achieve a batch size of 64 we use gradient accumulation (8 mini batches of batch size 8).

We implemented the intermediate fusion with short cuts, by pooling (average pool) feature embeddings from intermediate layers in the ResNet and PointVector backbones and concatenating them into the fully connected layers of the MLP fusion head.
The size of the hidden layers of the MLP being 1024, 1024, 512, and 256.
We used pre-trained weights for ResNet and PointVector, obtained through the single modality experiments described in Section~\ref{subsucsec:near-infrared-in-orthophotos} and ~\ref{subsec:3d-classification-performance-evaluation}.

The intermediate fusion approach is evaluated on the BioVista dataset and compared to the other tested 2D and 3D fusion approaches in Table~\ref{paper:g:tab:results-with-shortcut-fusion}. 
The 2D + 3D Intermediate Fusion with shortcuts approach, while hypothesized to benefit from features at multiple abstraction levels, did not yield performance improvements over other tested fusion techniques.
In terms of overall accuracy (79.7\% OAcc) and mean accuracy (79.8\% MAcc), it ranked lower than ensembling and both MLP fusion strategies. 
While its performance on the high forest biodiversity potential class was competitive with MLP fusion methods, it was surpassed by ensembling. 
For the low forest biodiversity potential class, it performed better than ensembling but was weaker than the MLP fusion approaches.
This approach did, however, still offer an improvement over using single modalities, which achieved OAccs of 76.7\% ± 2.6 and 75.8\% ± 0.4 for strictly 2D and 3D respectively. 
The observed suboptimal performance could be attributed to several factors. The use of average pooling to aggregate intermediate features might dilute crucial discriminative information. 
Exploring alternatives like max pooling could be beneficial.
Furthermore, the increased complexity from concatenating features from multiple layers might pose challenges for the MLP head to effectively learn optimal representations, potentially requiring more sophisticated architectural choices or regularization for the fusion module itself.

\begin{table}[th]
\footnotesize
\centering
\caption{Performance comparison of fusion approaches. We extend Table~\ref{tab:results} with the results from the intermediate fusion with short cuts to get an overview of the performance of all the tested fusion approaches. All metrics are means with associated standard deviation across seven runs with the same hyperparameter settings.}
\label{paper:g:tab:results-with-shortcut-fusion}
\begin{tabular}{@{}lrrrr@{}}
\toprule
\textbf{Model} & \multicolumn{1}{c}{\textbf{OAcc}} & \multicolumn{1}{c}{\textbf{MAcc}} & \multicolumn{1}{c}{$\textbf{Acc}_{\text{high}}$} & \multicolumn{1}{c}{$\textbf{Acc}_{\text{low}}$} \\\midrule
2D + 3D Ensembling & 80.5\% ± 2.0 & 80.6\% ± 1.9 & \textbf{89.5\%} ± 2.7 & 71.8\% ± 6.0 \\
\makecell[l]{2D + 3D MLP Fusion \\ (\replaced{frozen backbones}{pre-computed})} & \textbf{81.4\%} ± 0.5 &	\textbf{81.3\%} ± 0.5	& 85.8\% ± 4.5 & 76.9\% ± 3.8 \\
\makecell[l]{2D + 3D MLP Fusion \\ (end-to-end)} & 80.4\% ± 0.2  &	80.5\% ± 0.2	&  83.6\% ± 2.1 &	\textbf{77.3\% ± 2.1} \\
2D + 3D Intermediate Fusion & 79.7\% ± 1.2 & 79.8\% ± 1.1 & 86.2\% ± 3.5 & 73.4 \% ± 5.2 \\
\bottomrule
\end{tabular}
\end{table}
\end{appendices}

\newpage
\bibliographystyle{elsarticle-harv}
\bibliography{references} 

\newpage

\section{List of figure captions}
\listoffigures

\end{document}